\documentclass[11pt]{article}

\usepackage{amsmath}
\usepackage{amssymb}

\usepackage{cite}

\usepackage{hyperref}

\usepackage{lineno}

\usepackage{microtype}
\DisableLigatures[f]{encoding = *, family = * }

\usepackage{tabularx}

\usepackage[pdftex]{graphicx}

\usepackage[labelfont=bf,labelsep=period,justification=raggedright]{caption}

\makeatletter
\renewcommand{\@biblabel}[1]{\quad#1.}
\makeatother

\date{}

\begin{document}

\begin{flushleft}
{\Large
\textbf{The Filament Sensor for Near Real-Time Detection of Cytoskeletal Fiber Structures}
}
\\
Benjamin Eltzner$^{1,\ast}$,
Carina Wollnik$^{2}$,
Carsten Gottschlich$^{1}$,
Stephan Huckemann$^{1}$,
Florian Rehfeldt$^{2, \ast \ast}$
\\
{\bf 1} Institute for Mathematical Stochastics, Georg-August-University, 37077 G\"{o}ttingen, Germany
\\
{\bf 2} Third Institute of Physics -- Biophysics, Georg-August-University, 37077 G\"{o}ttingen, Germany
\\
$\ast$ Corresponding author email: benjamin.eltzner@mathematik.uni-goettingen.de\\
$\ast\ast$ Corresponding author email: rehfeldt@physik3.gwdg.de
\end{flushleft}

\section*{Abstract}

A reliable extraction of filament data from microscopic images is of high interest in the analysis of acto-myosin structures as early morphological markers in mechanically guided differentiation of human mesenchymal stem cells and the understanding of the underlying fiber arrangement processes. In this paper, we propose the filament sensor (FS), a fast and robust processing sequence which detects and records location, orientation, length and width for each single filament of an image, and thus allows for the above described analysis. The extraction of these features has previously not been possible with existing methods. We evaluate the performance of the proposed FS in terms of accuracy and speed in comparison to three existing methods with respect to their limited output. Further, we provide a benchmark dataset of real cell images along with filaments manually marked by a human expert as well as simulated benchmark images. The FS clearly outperforms existing methods in terms of computational runtime and filament extraction accuracy. The implementation of the FS and the benchmark database are available as open source.

\section*{Introduction}

In the last decade it has become evident that for cellular behavior the mechanical environment can be as important as traditionally investigated biochemical cues \cite{Discher2005,Rehfeldt2007}. Especially striking is the mechanically induced differentiation of human mesenchymal stem cells (hMSCs) cultured on elastic substrates of different elasticity \cite{EnglerSenSweeneyDisher2006}. During the early stage of this mechano-guided differentiation process in hMSCs, the structure and polarization of actin-myosin stress fibers depend critically on the Young's elastic modulus of the substrate  and can be used as early morphological markers \cite{Zemel2010a}. An analogous study showed that also for myoblast differentiation \cite{Yoshikawa2013}.

Stress fibers are contractile structures mainly composed of actin filaments, myosin motor mini filaments (non-muscle myosin II's) and distinct types of cross-linkers e.g. $\alpha$-actinin, fascin, etc.. These 'cellular muscles' are connected to the extra-cellular matrix via focal adhesions and generate and transmit forces to the outside world by pulling on the ECM proteins \cite{Wen2014}. Acto-myosin filaments are therefore considered key players in the mechano-sensing machinery of the cell that integrates physical cues from the surrounding to bio-chemical signaling and finally differentiation \cite{Swift2013}.

\begin{figure}[!ht]
  \includegraphics[width=\textwidth]{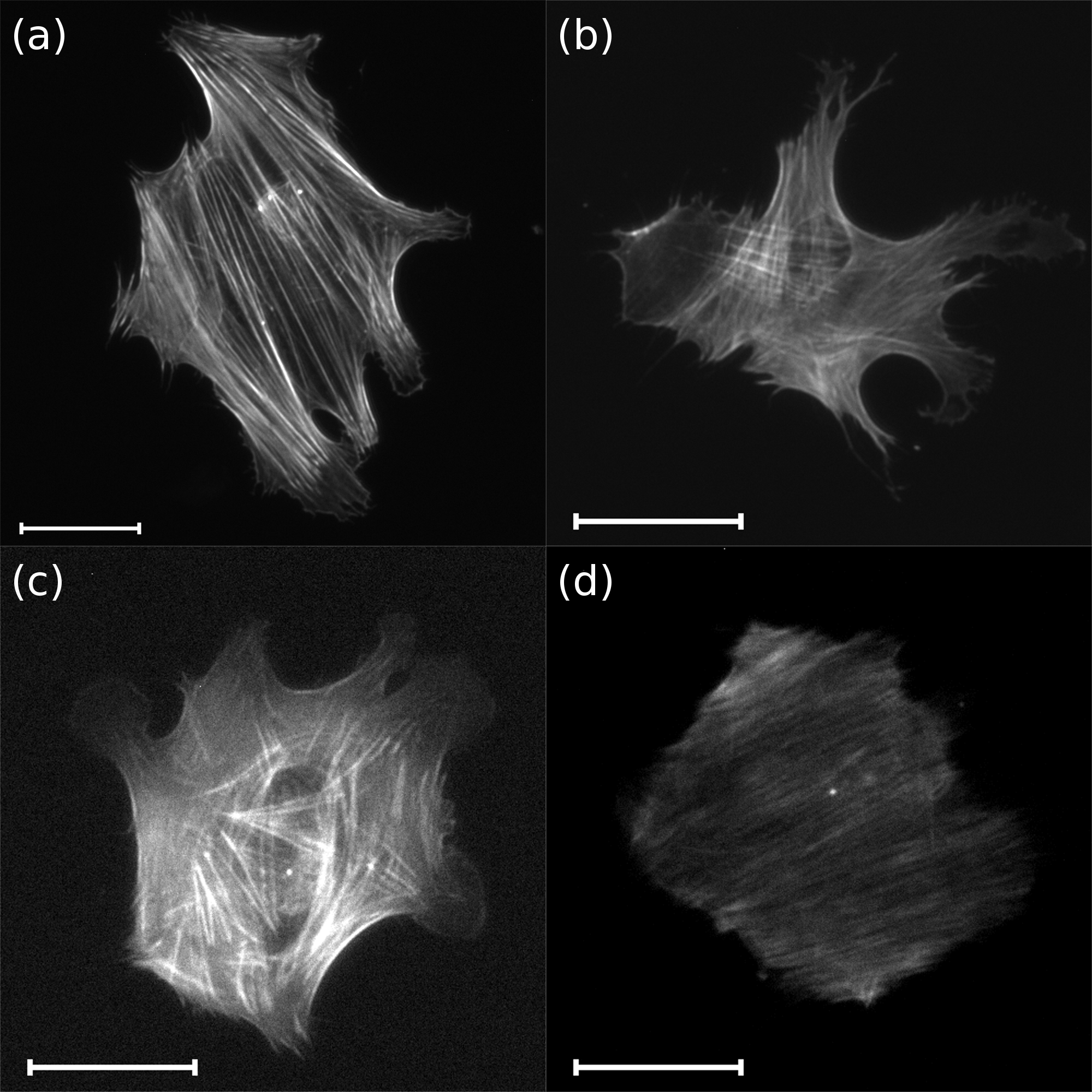}
  \caption{{\bf Varying quality images of human mesenchymal stem cells in view of filament expression.} Subfigures: (a)~\textit{good~quality} image G1 of an immunoflurescently stained fixed cell of large size with clearly visible stress fibers on gel with stiffness 10 kPa; (b)~\textit{medium~quality} image M3 of an immunoflurescently stained fixed cell of moderate size with inhomogeneous brightness and slight blur on glass; (c)~\textit{poor~quality} image B2 of a live cell of moderate size with considerable noise and excessive brightness due to overexposure on glass; (d)~\textit{very~poor~quality} image VB2 of a live cell of moderate size with very low contrast due to bleaching, considerable blur and hardly discernible stress fibers.}
  \label{image_examples}
\end{figure}

To further elucidate and potentially model the complex mechanical interplay between matrix and the cell a full description and understanding of the structure and dynamics of acto-myosin stress fibers is essential. Using fluorescence microscopy, filament arrangements can be visualized where stress fibers can be considered as linear filaments of varying width and length. Typical images of acto-myosin stress fibers of different quality are displayed in Fig. \ref{image_examples}. This poses certain challenges for a reliable extraction of all filaments' location, length, width and orientation (e.g. relative to the cell's major axis). In particular, studies of the dynamics of stress fibers where multiple time series of living cells are recorded in parallel over periods of 24 hours, demand a fast and reliable fiber detection algorithm. Such experiments are essential towards a more detailed understanding of the role of acto-myosin cytoskeleton structure formation during the mechanically induced differentiation of hMSCs \cite{Zemel2010a}.

While our motivation is based on the cytoskeleton structure of hMSCs, there is also demand to trace and track stress fibers over space and time in other research areas. For example, studies on migrating cells indicate \cite{HotulainenLappalainen2006, Naumanen2008, PellegrinMellor2007, Ciobanasu2012, Tojkander2012, Vallenius2013} various stress fiber types (dorsal, ventral, arc) appearing at different locations inside a migrating cell. Their exact cellular function yet remains unknown but could be clarified by live cell imaging. Following the filament dynamics over time gives further insight into the formation and function of stress fibers. Recently, Soine et al. described a novel method to analyze traction force microscopy data, so called model-based traction force microscopy \cite{Soine2015}. Here, it is imperative to detect and mark the stress fibers of a cell to gain more insight into cellular force generation and transmission to the substrate. Such live cell experiments are ideally performed for many cells in parallel to achieve sufficiently significant statistics, therefore the fiber analysis algorithm ideally performs tracing and tracking in (nearly) real-time.

With the Filament Sensor we provide such an image processing tool that yields the stress fiber structure from observations of live as well as fixed cells in terms of images with widely varying brightness, contrast, sharpness and homogeneity of fluorescence, cf. Fig. \ref{image_examples}. In our live cell imaging setup, typically 30 cells are followed over a period of 24 hours with an image taken every 10 minutes. Thus, the aim of real time processing allows for about 20 seconds of processing time per image.

Of course, such a tool can be employed to extract filament features for any (sets of) images containing fiber structures. Applications are conceivable in a wide range of cases especially in the context of actin fiber structures, e.g. \cite{Tojkander2012,Sanchez2012}, but also more generally in biology, medical imaging and material science.

\subsection*{The Filament Sensor and the Benchmark Dataset}\label{challenges:subscn}

To obtain the full information set of the stress fibers in cells, namely location, length, width, and orientation, from repeated observations of living cells under widely varying conditions in near real time demands from the \emph{filament sensor} (FS) to extract

\begin{figure}[!ht]
  \includegraphics[width=\textwidth]{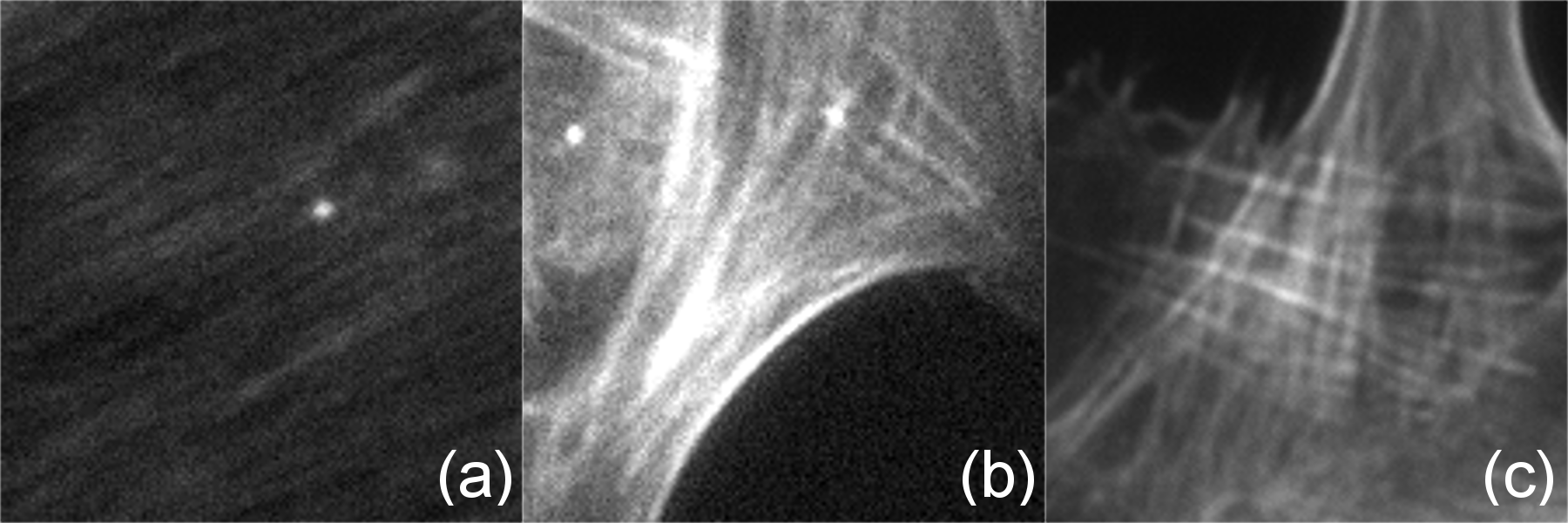}
  \caption{{\bf Challenges for filament extraction.} (a)~blur~(detail~from~image~VB2) The overall contrast of the cell body is very low and lines are hardly discernible. (b)~overexposure~and~noise~(B2) The extensive regions of maximal brightness hide any structure that may be present in those regions. Salt and pepper noise is clearly visible as dark spots in bright areas and bright spots in dark areas. (c)~filament~crossings~(M3). A bundle of roughly vertical filaments of varying brightness crosses a bundle of roughly horizontal filaments with varying brightness.}
  \label{image_challenges}
\end{figure}

\begin{enumerate} 
  \item[I)] fast and unsupervised
  \item[II)] robustly 
  \item[III)] all filament features: location, length, width and orientation;
\end{enumerate}
where II) implies dealing with several specific problems illustrated in Fig. \ref{image_challenges}
\begin{enumerate} 
  \item[IIa)] detecting darker lines crossing bright lines,
  \item[IIb)] dealing with image inhomogeneities and
  \item[IIc)] dealing with image blur and noise.
\end{enumerate}

\begin{figure}[!ht]
  \includegraphics[width=\textwidth]{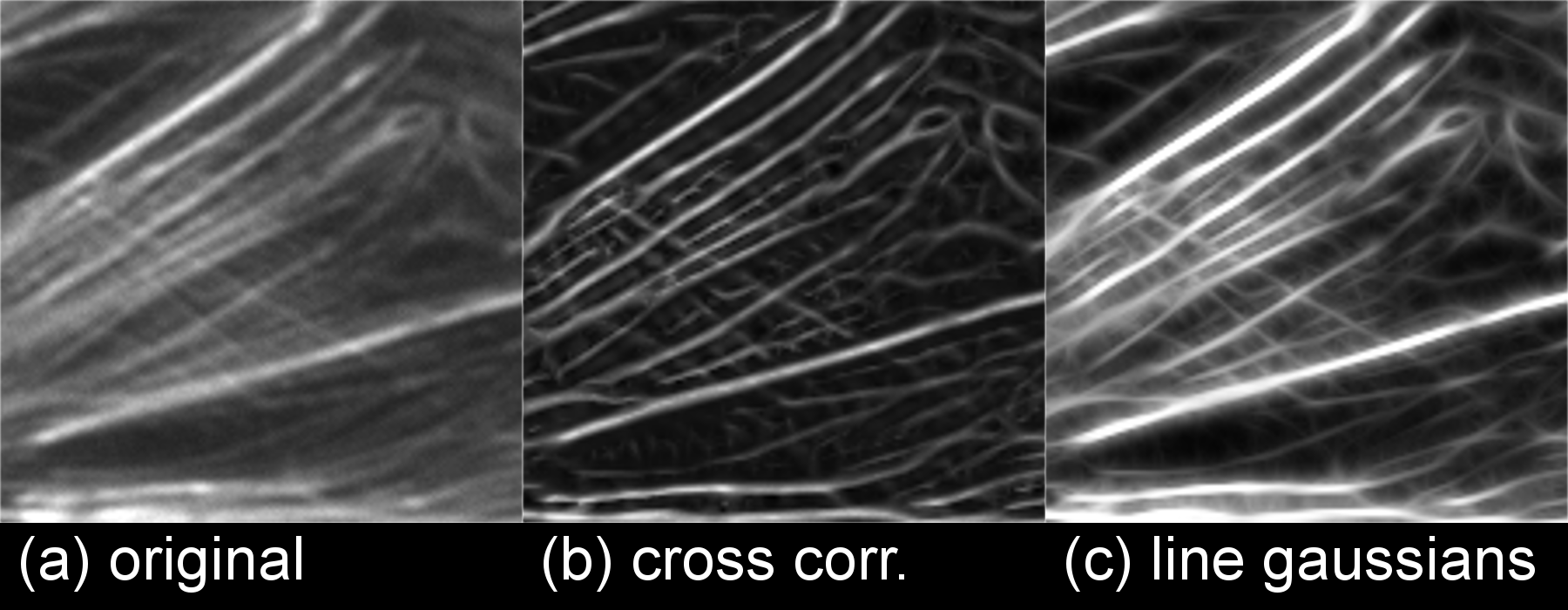}
  \caption{{\bf The cross correlation filter suppresses faint lines.} (a)~original. The lines in question are discernible to the human eye but contrast is low. (b)~normalized~cross~correlation. Contrast is greatly enhanced but faint lines are severed by crossing bright lines (c)~line~Gaussians. Contrast and overall brightness are enhanced. Especially, faint lines crossing bright lines are not suppressed. A line illustration these findings particularly well is running from the upper left to the lower right.}
  \label{line_severed}
\end{figure}

The FS is specifically designed to meet these challenges. Dealing with image inhomogeneity calls for the application of local image processing tools. Blurring effects will be mitigated by line enhancement through direction sensitive methods. Crossings of lines of varying intensities can be rather successfully detected by what we call \emph{line Gaussians} which are specific oriented thin masks, cf. Fig. \ref{line_severed}. After local binarization, finally an adaption of the semilocal line sensor approach to fingerprint analysis \cite{GottschlichMihailescuMunk2009} is applied to extract all filament features. As the FS is modularized, employs local and orientation dependent image analysis methods and outputs the entire filament data, expert knowledge such as detecting fewer filaments in specific low variance areas, say, can be easily incorporated.

To assess our method we have devised two benchmark datasets. One set comprises filaments in hMSCs of varying image quality, manually labeled by an expert. The second database consists of simulated fiber structures providing a test scenario with complete knowledge of spatial information.

In order to compare our new method to existing methods (discussed in Section ``Results'') we have to restrict our comprehensive output to the limited output of others. Specifically for eLoG (elongated Laplacian of Gaussian) method \cite{Zemel2010a} as well as for the Hough transformation such limited output is given by angular histograms: accumulated pixel length of filaments per angular orientation. For the constrained inverse diffusion (CID) method \cite{BasuDahlRohde13} we can only compare sets of pixels which is also possible for the eLoG method. 

A Java implementation of the FS, the benchmark datasets and ground truth data, and a python script for evaluation are available under free open source and open data licenses under the project's web page
\url{http://www.stochastik.math.uni-goettingen.de/SFB755\_B8} .

\section*{Related Work}

Despite the existence of a tremendous body of techniques for image processing and especially line detection (for an overview e.g. \cite{Szeliski2010}, Chapter 4), currently the methods used for filament identification in cell images are often ad hoc, require partial manual processing and a fair amount of runtime, e.g. \cite{Zemel2010a,FaustHampe2011,BasuDahlRohde13}. The latter two requirements are particularly undesirable if large numbers of images are to be evaluated, as is typically the case.

Fundamental global methods in this context include the Hough transformation \cite{DudaHart1972} and brightness thresholding via the Otsu method \cite{Otsu1975}. However, variable brightness of cell plasma and filaments demand that such methods must at least be supplemented by local methods like the Laplacian of Gaussian (LoG, e.g. \cite{Marr80,HaralockShapiro1991}), anisotropic diffusion (e.g. \cite{Weickert1996,GottschlichSchoenlieb2012}), or a beamlet approach, e.g. \cite{Donoho2002}.

Many algorithms exist that focus on the analysis of networks of strongly curved microtubuli (as opposed to the properties of single filaments), such as line thinning \cite{Chang2001}, active contours \cite{Dormann2002} and the recently proposed constrained inverse diffusion (CID) method \cite{BasuDahlRohde13}.

Other methods which are geared toward extraction not only of filament pixel position but also of local orientation include the FiberScore algorithm \cite{Lichtenstein2003}, elongated Laplacians of Gaussians (eLoGs) \cite{Zemel2010a} and gradient based methods \cite{Herberich2010,FaustHampe2011}. 

The eLoG method, like the gradient method aims at detecting not only filament pixels but also
their orientations. While filament length and width are not extracted, counting the number of pixels per orientation, these methods yield histograms of cumulated filament length per orientation angle which are then further analyzed \cite{Zemel2010a,FaustHampe2011}.

Line thinning and CID identify only a skeletal filament network structure without filament orientation, length, nor width. 
Rather, these methods are geared towards detection of thin microfilaments and not of wide stress fibers we are interested in. 

The CID method uses a maximum cross correlation coefficient method for templates with different orientations as a preprocessing step. While this yields considerable line enhancement in many cases, in some cases it leads to situations where bright lines sever fainter lines they cross, as illustrated in Fig. \ref{line_severed}. Certainly, this is undesirable in view of the aforementioned challenge IIa. One could try to account for this by adding cross masks to the list of filters. This, however, increases the number of filters and thereby calculation time considerably; additionally bright lines acquire a halo.

The FiberScore program \cite{Lichtenstein2003} produces local orientation and centerline images and provides global information on accumulated line length and average width. It produces no line objects as output. It turned out that the methods applied in FiberScore did not yield optimal results for our cell images. The most important obstacle, however, for using FiberScore is that neither the program's nor its framework's source codes are freely available. Although the original developer has very helpfully supported us to make the program run we could not tailor FiberScore to our needs.


An overview over the different types of output of some methods is given in Table \ref{table}.

\begin{table}[!ht]
  \begin{center}
    \begin{tabular}{| >{\raggedright\let\newline\\}m{3.2cm} |*{3}{>{\centering}m{1.1cm}|} >{\centering\arraybackslash}m{2cm}| >{\centering\arraybackslash}m{1.1cm}| }
      \hline
                                         & CID        & eLoG       & Hough      & FiberScore & FS         \\ \hline
      \vspace*{0.5\baselineskip}
      line pixels
      \vspace*{0.5\baselineskip}         & \checkmark & \checkmark &            & (\checkmark) & \checkmark \\ \hline
      \vspace*{0.5\baselineskip}
      angular histogram
      \vspace*{0.5\baselineskip}         &            & \checkmark & \checkmark & (\checkmark) & \checkmark \\ \hline
      pixelwise orientations             &            & \checkmark &            & (\checkmark) & \checkmark \\ \hline
      overall line length and width      &            &            &            & (\checkmark) & \checkmark \\ \hline
      individual line length and width   &            &            &            &            & \checkmark \\ \hline
      \vspace*{0.5\baselineskip}
      runtime
      \vspace*{0.5\baselineskip}         & $\sim 20h$ & $\sim 20m$ & $\sim 20s$ & $\sim 20s$ & $\sim 20s$ \\ \hline
    \end{tabular}
  \end{center}
  \caption{Data collected by various methods and comparison of runtime. Checkmarks in parentheses indicate that for lack of documentation and source code availability the FibreScore method is not portable in terms of full configurability and usability to a generic system.}
  \label{table}
\end{table}

\section*{Materials and Methods}

The cells used in this work are human mesenchymal stem cells derived from human bone marrow, acquired from the commercial vendor Lonza \cite{Lonza} under the product number PT-2501. The vendor anonymizes personal data of cell donors.

\begin{figure}[!ht]
  \centering
  \includegraphics[height=0.85\textheight]{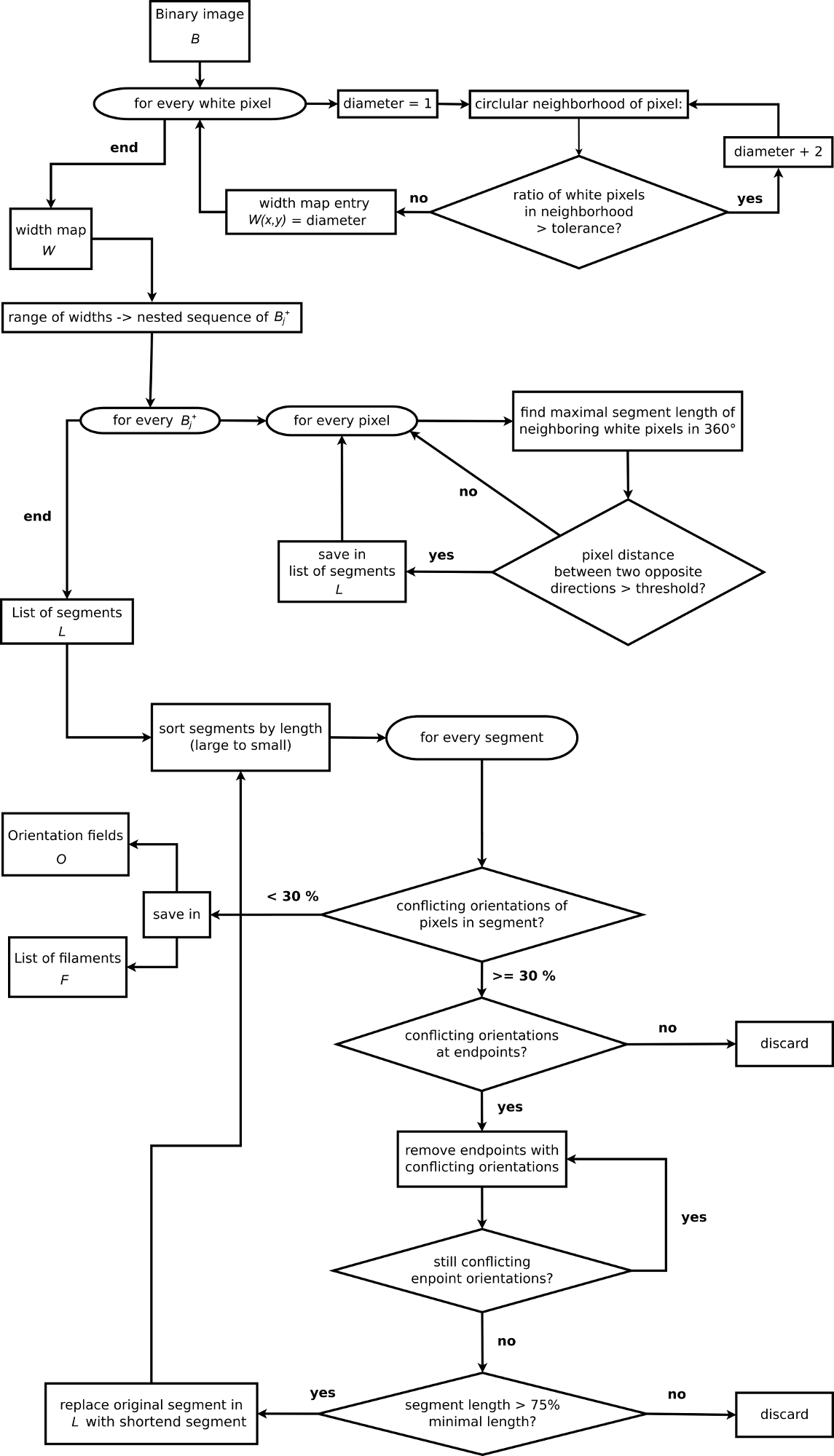}
  \caption{{\bf Flow chart illustrating the algorithm of the line segment sensor.} The algorithm begins at the very top with a binary image and outputs an orientation field and line information, displayed to the bottom left of the chart. A detailed explanation is given in the text.}
  \label{flow_chart}
\end{figure}

Our proposed method decomposes into two parts. First, a set of specifically tailored image enhancement and binarization procedures are applied, followed, secondly, by a width aware segment sensor, which extracts filament data from the binary image. For this core part of the FS its involved workflow is sketched in Fig. \ref{flow_chart}. Suitable default values detailed below have been determined by expert knowledge on hMSCs images from a database separate from the two databases we have tested on in Section ``Results''.

\subsection*{Parameter Choice and Typical Workflow}

The empirically determined default values proposed by the FS yield good results for most images. Otherwise, starting from the default values a local optimum in the parameter space can usually be determined very quickly. Tracing results are fairly weakly dependent on parameter variation in the optimal parameter region and image properties do not vary strongly over individual 24-48 hours time series of images with overall acceptable quality. Therefore, to commonly adjust parameters for a whole batch of images, it is usually sufficient to consider one of the first and one of the last images only. In this sense, the FS is ``semi-automated''.

\subsection*{Preprocessing and Binarization}

Apart from an adjustable contrast enhancement and a normalization of the image to 256 gray levels, only local methods are employed. The first two preprocessing steps use generic Gaussian and Laplacian filters with adjustable variance and magnitude (with empirically successful default values), respectively, the order of which can be interchanged. 

\begin{figure}[!ht]
  \includegraphics[width=\textwidth]{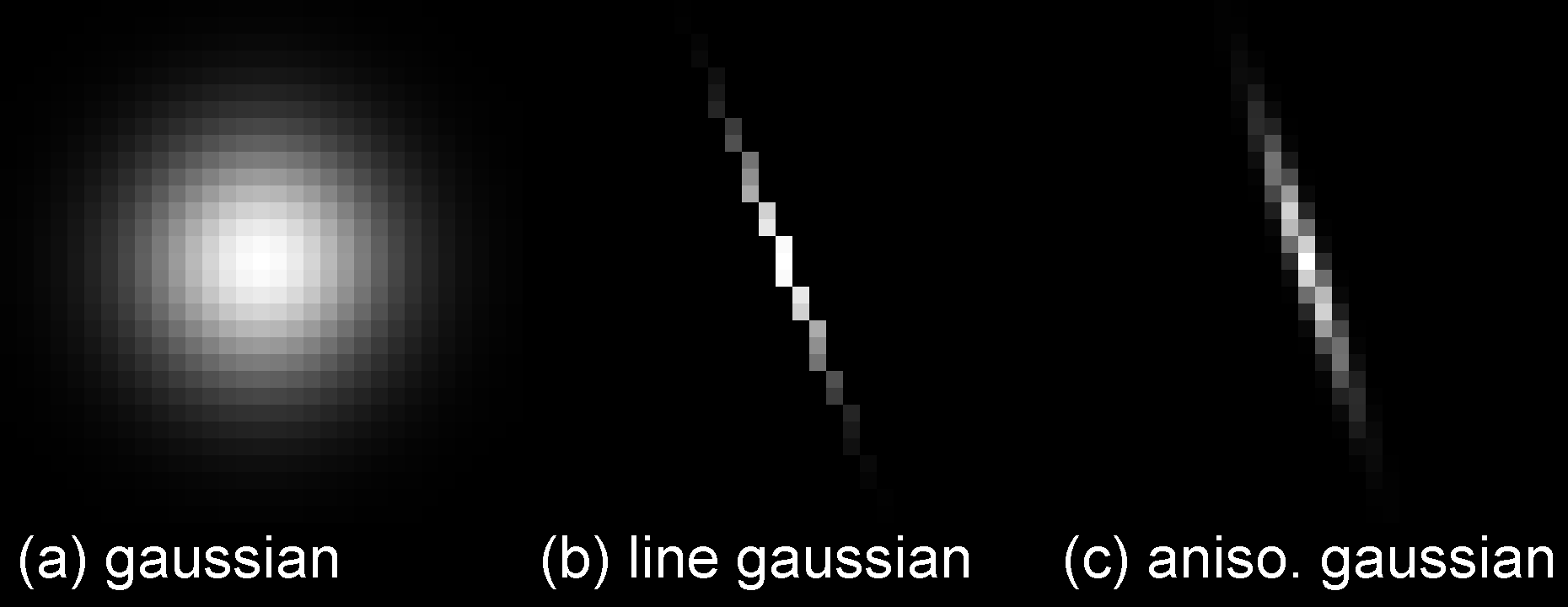}
  \caption{{\bf Illustration of the line Gaussian.} (a) an isotropic two dimensional Gaussian. Convolving the image with such a filter will lead to blurring of the image. (b) restriction of an isotropic Gaussian to a line. This filter locally homogenizes pixel brightness along lines that run in direction of the filter. It is computationally efficient as it only uses few pixels. (c) an elongated Gaussian. This filter also homogenizes lines, but it has much more Pixels and thus requires much longer computation.}
  \label{line_gauss}
\end{figure}

\paragraph{The linear Gaussian.} In the third preprocessing step we accumulate intensities along rod kernels which have been used for cross correlations \cite{Lichtenstein2003}. This filter is orientation sensitive and thus very well suited for line enhancement. It is comparable to the eLoG approach but performs much better in terms of calculation time because the filter uses a Gaussian mask which is then restricted to lines of different orientations only and convolved with the image, cf. Fig. \ref{line_gauss}. In our implementation we choose the filter size to be $l = 2 \cdot \lceil 3 \sigma\rceil + 1$ pixels (here ``$\lceil\cdot\rceil$'' denotes rounding to the next integer and $\sigma$ is $3$ pixels by default) and we use $2l-2$ lines; one line per pair of opposite boundary pixels of the mask. Because the filter operation is only performed on pixels on a line, it is by a factor of approximately $l$ faster than the eLoG approach which uses square filter masks. The highest response is taken as the new grey value of the pixel. In fact, this filter proves to be very successful at enhancing darker lines crossing bright lines, cf. Fig. \ref{line_severed}.

\paragraph{Local binarization} is achieved via a combination of Gaussian weighted adaptive means \cite{GonzalesWood2001} and global thresholding, where the global threshold is not compared to single pixels but to neighborhood means. The global threshold is usually set to only about $10\%$ of the maximum brightness and serves mainly to reduce calculation time by removing obvious background from consideration.

\paragraph{Homogeneous noise cancellation} is achieved by one of two filters. The default filter calculates mean brightness along rods as used in the linear Gaussian filter around the white pixels of the binary image and then thresholds the ratio of standard deviation to a function $f$ of the mean $\mu$ of these mean brightnesses. The simpler alternative filter thresholds the ratio of Gaussian weighted standard deviation and the same function $f$ of the mean $\mu$ of pixel brightnesses in neighborhood of white pixels. The function $f$ is given by
\begin{align*}
  f(\mu) = \left\{\begin{array}{lcl} \frac{4}{9} (\mu - 36 \sqrt{\mu} + 100) & \text{for} & \mu>30\\ \mu & \text{else} &\\ \end{array} \right.
\end{align*}

Optionally, a filter performing morphological ``closing'' (cf. \cite{Szeliski2010}, Chapter 3.3.2) can be applied to the binary image.

\subsection*{A Width Aware Segment Sensor}

After preprocessing and binarization, filament data is now extracted from the white pixels using the following algorithm, illustrated by a flow chart in Fig. \ref{flow_chart}. Denote the binary image by $B$ and its pixel value at position $(x,y)$ by $B(x,y)$ and denote the subset of white pixels by $B^+$
\begin{enumerate}
  \item Every white pixel $B(x,y) \in B^+$ is assigned a \emph{width}, $W(x,y)$ which yields a width map $W$. This is done by taking circular neighborhoods of the pixel (cf. Fig. \ref{circle_mask}) with increasing diameter. A diameter is accepted, if the ratio of white pixels of the binary image is above an adjustable tolerance (with default value $95\,\%$). If a diameter was accepted, the next larger diameter is tested until a diameter is rejected. The width $W(x,y)$ at the pixel is then given by the largest accepted diameter at the pixel. In particular, this gives a range of widths $1\geq w_1 < \ldots < w_k = \max W(x,y)$ attained by pixels in $W$ and it extends the binary image $B$ to a nested sequence of binary images $B_j$ with white pixels $B_j^+ = \{(x,y)\in W: W(x,y) \geq w_j\}$, $j=1,\ldots,k$. A temporary List $L$, the \emph{filament data set} $\mathcal{F}$, and the \emph{orientation field} $\mathcal{O}$ are each initialized by the empty set.

  \begin{figure}[!ht]
    \includegraphics[width=\textwidth]{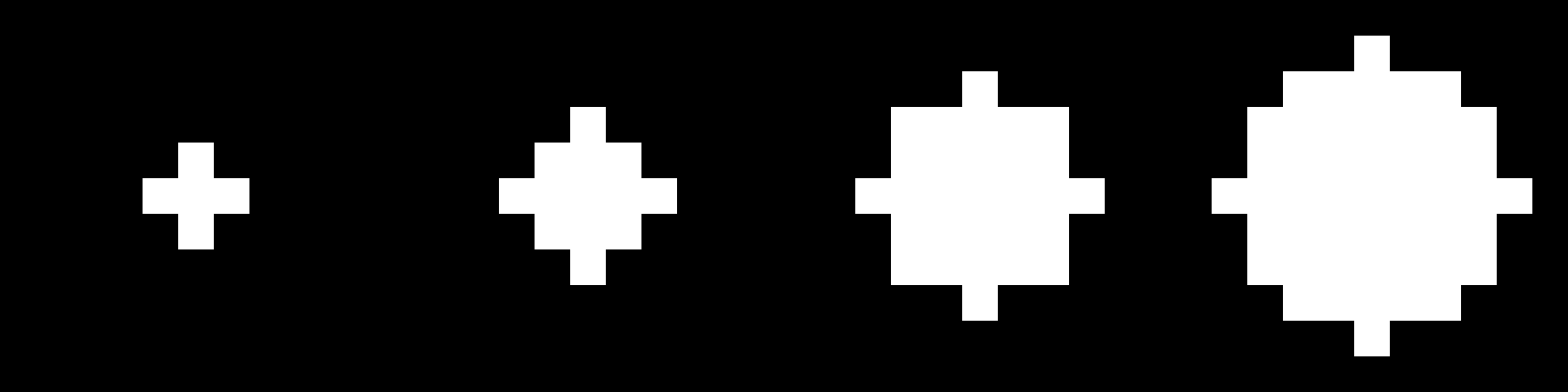}
    \caption{{\bf Some circle masks.} These are examples of the circular masks used by the segment sensor algorithm to determine line width. The circles displayed here correspond to diameters of $2$, $4$, $6$ and $8$ pixels. The masks are squares with an odd number of pixels as they are centered at a unique pixel.}
    \label{circle_mask}
  \end{figure}
  
  \item For every $j = k,\ldots, 1$ in decreasing order, apply the \emph{segment sensor} to $B_j^+$ as follows.
  \begin{enumerate}
    \item For each pixel $B_j(x,y) \in B_j^+$ the segment sensor probes into a number of directions in $B_j$ ($1^\circ,\ldots,360^\circ$ by default; this corresponds to $180$ orientations). For each direction it determines the maximal length at which pixels can be found, connected by a straight line to $B_j(x,y)$ in $B_j^+$ as illustrated in Fig. \ref{spokes}. The pixel data of the largest line segment acquired as the union of the lines of two opposing directions is stored to $L$, if its length exceeds an adjustable threshold of minimal filament length ($20$ pixels by default). These pixel data only include the centerline pixels found in $B_j^+$.

    \begin{figure}[!ht]
      \includegraphics[width=\textwidth]{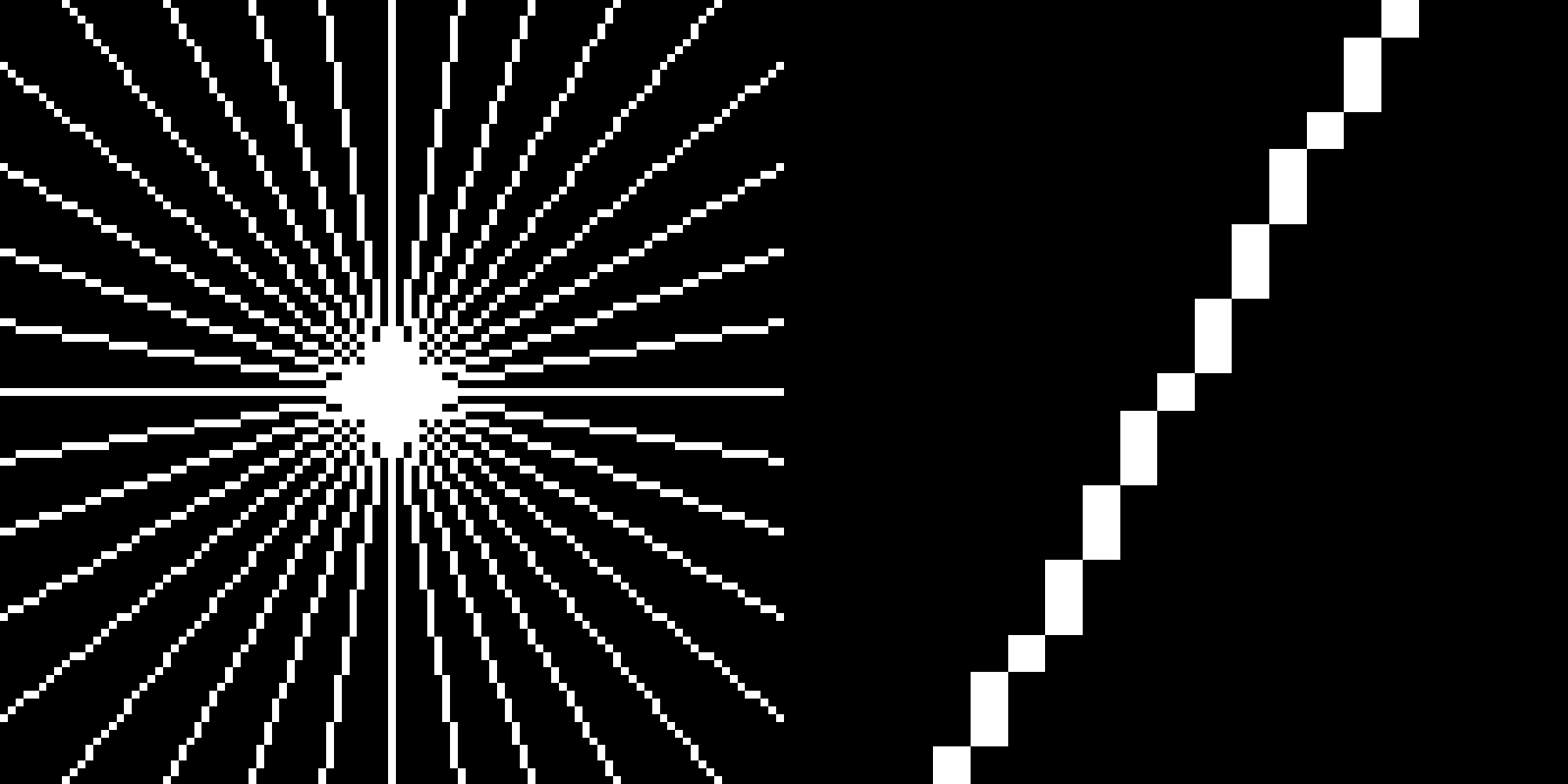}
      \caption{{\bf Illustration of the segment sensor's probing paths.} Some connectivity lines (left) with detail (right) along which the segment sensor probes. The lines illustrated correspond to multiples of $10^\circ$ to render the illustration legible. The algorithm uses multiples of $1^\circ$, which implies ten times as many paths. The detail shows that the lines are thin in the sense that pixels on diagonal lines touch only at the corner points.}
      \label{spokes}
    \end{figure}

    \item In the next step, segments in $L$ are called in the order of their length, long segments first. For every segment, the orientation field $\mathcal{O}$ (which is empty when first called) is looked up for every pixel on the segment. If less than $30\,\%$ of its pixels have a \emph{conflicting orientation} entry in $\mathcal{O}$, -- i.e. the entry in $\mathcal{O}$ differs by less than an adjustable minimal tolerance angle (of default value $20$ degrees) from the segment's orientation -- the segment is accepted as valid. For every pixel within a circular neighborhood with diameter $w_j + 2$ pixels (in order to avoid duplication) of a segment pixel, the segment's orientation is stored to $\mathcal{O}$, if $\mathcal{O}$ does not yet have an entry there. The segment is then also added to $\mathcal{F}$. If at least $30\,\%$ of the pixels on a segment have a conflicting orientation, we have the following cases.
    \begin{enumerate}
      \item If $\mathcal{O}$ does not carry a conflicting orientation for any of the endpoints, the segment is discarded.
      \item Otherwise, the endpoints with conflicting orientations are iteratively removed from the segment until the remaining segment's endpoints no longer have a conflicting orientation. If the resultant segment length is above 75\% of the threshold of minimal filament length, this new segment is added back to $L$ and the original one is removed. The new segment is revisited when its length is called.
    \end{enumerate}
  \end{enumerate}
\end{enumerate}

As lines are blurred due to scattering and as the preprocessing usually enhances line width, the FS tends to find greater line width than a human expert. Taking into account this expert knowledge, the FS returns a filament width reduced by one for filaments with width larger than $1$. Such and other adjustments are feasible for the FS because it extracts filaments individually.

\subsection*{Benchmark Datasets}

To obtain benchmark images of different quality, images of cells that were bleached by long exposure to light, less bleached cells from time series with moderate exposure to light and completely unbleached fixed cells have been chosen. From this pool, 10 images, two of very poor quality (labeled $VB1$ and $VB2$ below), two of poor quality ($B1$ and $B2$), three of medium quality ($M1$, $M2$,$M3$), and three of good quality ($G1$,$G2$, $G3$) have been selected and their stress filament structure manually labeled by an expert.

\paragraph{A filament quality score.} Measures for general quality of images are abundant; however, here, we seek for a quality measure specifically tailored to filament images. Our use case is related to the context of fingerprint analysis, where images with curved, linear and highly parallel structures are investigated. Here we propose a quality measure that is motivated from the usage of Gabor filter variance for quality measurement in fingerprint analysis from \cite{ShenKotKoo2001}. 

\begin{figure}[!ht]
  \includegraphics[width=\textwidth]{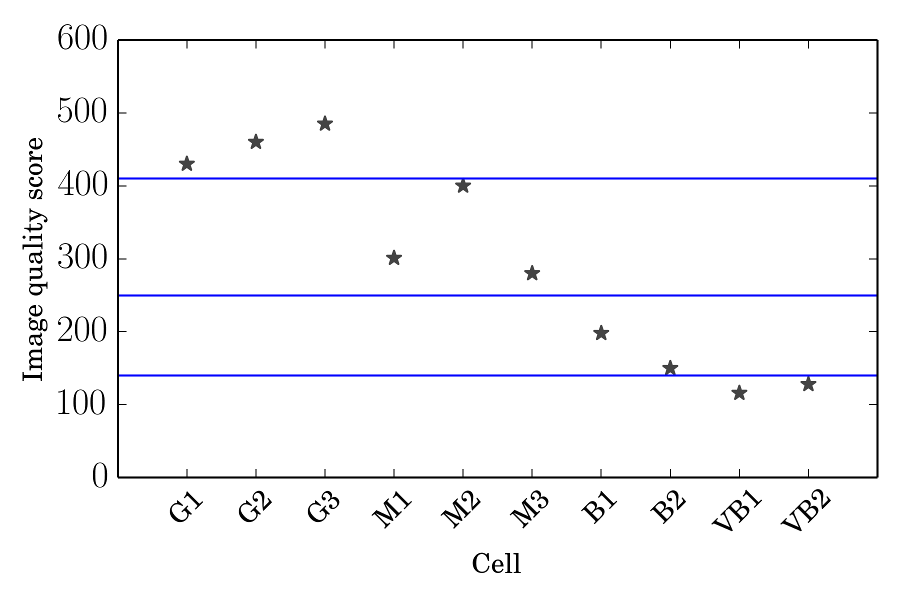}
  \caption{{\bf Filament quality scores of benchmark database images.} FQS is determined in terms of line sharpness and contrast, size of the cell, and bright spots due to overexposure, where the latter decreases the score, while the line sharpness and contrast and cell size contribute positively. The blue lines indicate our separation of the images into qualitative classes. The classes were chosen to contain like numbers of images and only serve illustrative purposes as a rough quality label.}
  \label{quality_scores}
\end{figure}

Define the \emph{filament quality score} as
\begin{align}\label{FIQS}
  FQS = 1500 \frac{N_o}{N_a} + 0.001 N_a - 7500 \frac{N_b}{N_a}
\end{align}
where $N_o$ is the number of pixels $(x,y)$ such that
\begin{align*}
  3 \sqrt{2 \pi} \max_{\text{rod}} \sum_{(d_x, d_y) \in \text{rod}} \limits G_3(d_x, d_y) I(x+d_x, y+d_y) > 1.1 \sum_{d_x = -9}^{9} \sum_{d_y = -9}^{9} \limits G_3(d_x, d_y) I(x+d_x, y+d_y)\,,
\end{align*}
$N_a$ is the cell area measured in terms of the number of pixels and $N_b$ be the number of pixels $(x,y)$ such that
\begin{align*}
  \sum_{d_x = -9}^{9} \sum_{d_y = -9}^{9} \limits G_3(d_x, d_y) I(x+d_x, y+d_y) > 200\,.
\end{align*}
Here $I$ denotes the image to be tested and $G_3$ a Gaussian mask with standard deviation $\sigma = 3$. The resulting FQSs as calculated by the FS are displayed in Fig. \ref{quality_scores}. The constants in \eqref{FIQS} have been chosen such that the three terms' contributions are of the same magnitude. The optimization of the FQS is beyond the scope of this paper and the subject of separate research.

The FQS thus determines image quality in terms of three features.
\begin{enumerate}
  \item Sharpness and contrast of lines are determined using the means of Gaussians along rods and comparing the highest response to a Gaussian weighted neighborhood mean. The relative number of cell pixels where this ratio exceeds a threshold contribute to the score. This criterion is inspired by \cite{ShenKotKoo2001}.
  \item The size of the cell in the image contributes positively to the score as a small cell area leads to a high boundary to bulk ratio and a small number of lines, both of which are error sources.
  \item Marked bright spots due to overexposure contribute negatively to the score. As we are looking for bright structures, high brightness is more problematic than low brightness, because it acts as a smearing effect on lines. Therefore, we do not penalize low brightness regions likewise.
\end{enumerate}

For simulations of another 10 images we take a simple straightforward approach. The filament process is viewed as a random marked Poisson point process (e.g. \cite{BenesRataj2004}) where the marks indicate length and orientation of filaments centered at the respective points. Only filaments with their center point contained in an independently generated ellipse are recorded (unbiased sampling) and the cytoplasm background is mimicked by strongly blurring a subset of the filament pixels. We let angular orientation follow an independent mixture of one or two wrapped Gaussians -- giving an angular distribution with a realistic mode pattern \cite{HuKimMunkRehSommWeickWoll2014}. Grey levels of cell background and of filaments are independently perturbed; the resulting image is blurred by a Gaussian filter, and finally, independent Gaussian white noise is added. 

\begin{figure}[!ht]
  \includegraphics[width=\textwidth]{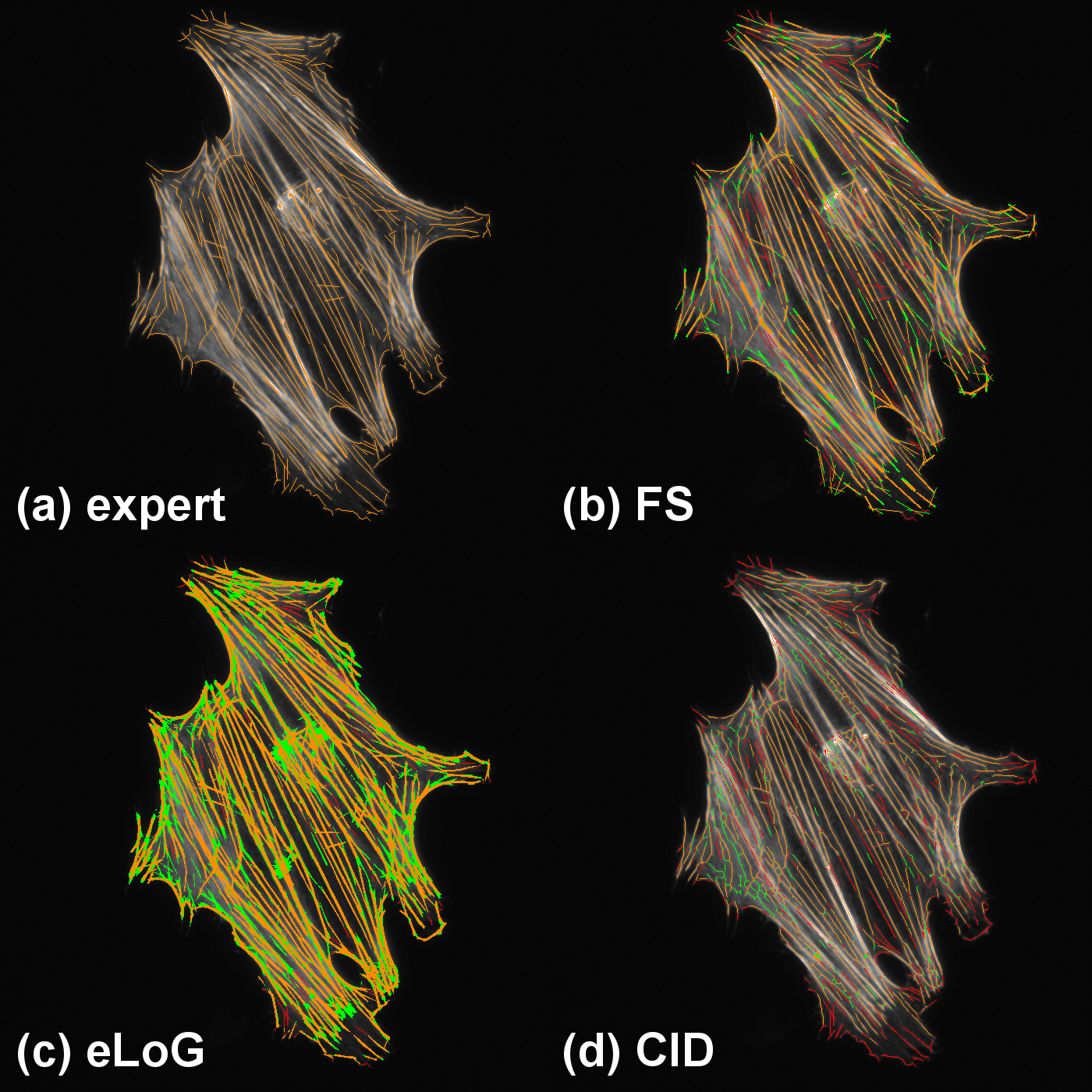}
  \caption{{\bf Tracing results with false positives and missed filaments for an hMSC.} The cell G1 used for this illustration is a fixed cell that has been immunofluorescently stained. The subfigures display: (a)~ground~truth, (b)~FS~results, (c)~eLoG~results, (d)~CID~results. Green pixels are false positives detected by the method, yellow are correctly identified pixels and red are missed pixels. A pixel is correctly identified, if it corresponds to a ground truth labeled pixel within an 8-neighborhood. A ground truth labeled pixel is considered missing, if no pixel was detected within an 8-neighborhood.}
  \label{pixels_image_real}
\end{figure}

\begin{figure}[!ht]
  \includegraphics[width=\textwidth]{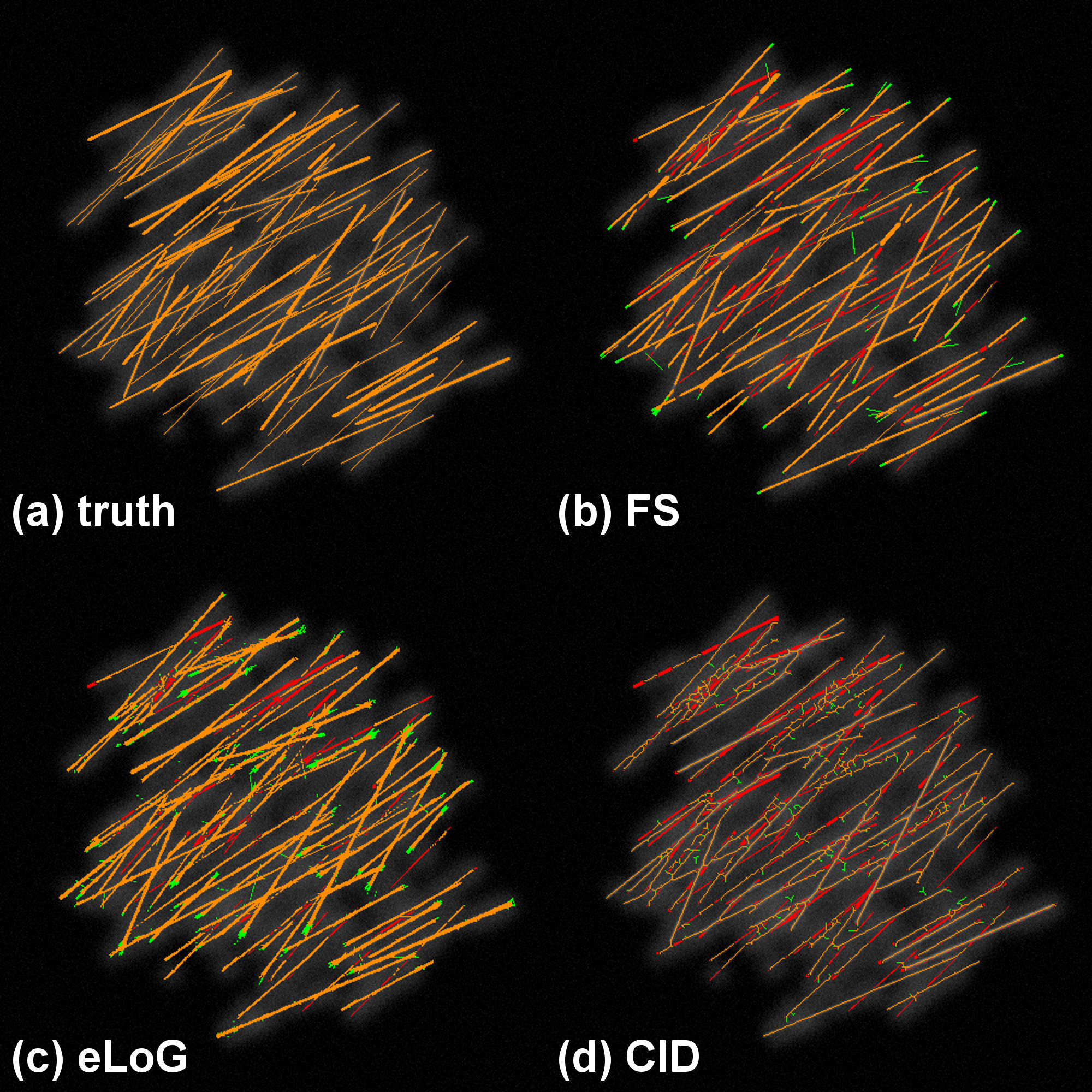}
  \caption{{\bf Tracing results with false positives and missed filaments for a simulated cell.} This illustration uses the simulated cell 05. Green pixels are false positives detected by the method, yellow are correctly identified pixels and red are missed pixels as in Fig. \ref{pixels_image_real}.}
  \label{pixels_image_sim}
\end{figure}

Like the FS, also eLoG and CID allow for visual display of filament pixels. Along with the ground truth we provide such images in Fig. \ref{pixels_image_real} for a real cell image as well as in Fig. \ref{pixels_image_sim} for a simulated cell image.

\section*{Results}\label{results:scn}

The performance of the FS is assessed via two benchmark datasets and compared against three existing methods for which implementations are available.
\begin{description}
 \item[eLoG] (elongated Laplacian of Gaussian) where we rely on \cite{Zemel2010a} but use a faster implementation of our own,
 \item[HT] (Hough transformation) following \cite{DudaHart1972}, and
 \item[CID] (constrained inverse diffusion) by \cite{BasuDahlRohde13}.
\end{description}
Since none of these methods provides for the full filament data, as elaborated above, comparison based on ground truth is only possible via \emph{angular histograms} recording accumulated length (for eloG and HT) or via \emph{centerline pixel detection} (for CID and eLoG) detailed below.

\subsection*{Qualitative Comparison}

\begin{figure}[!ht]
  \centering
  \includegraphics[height=0.8\textheight]{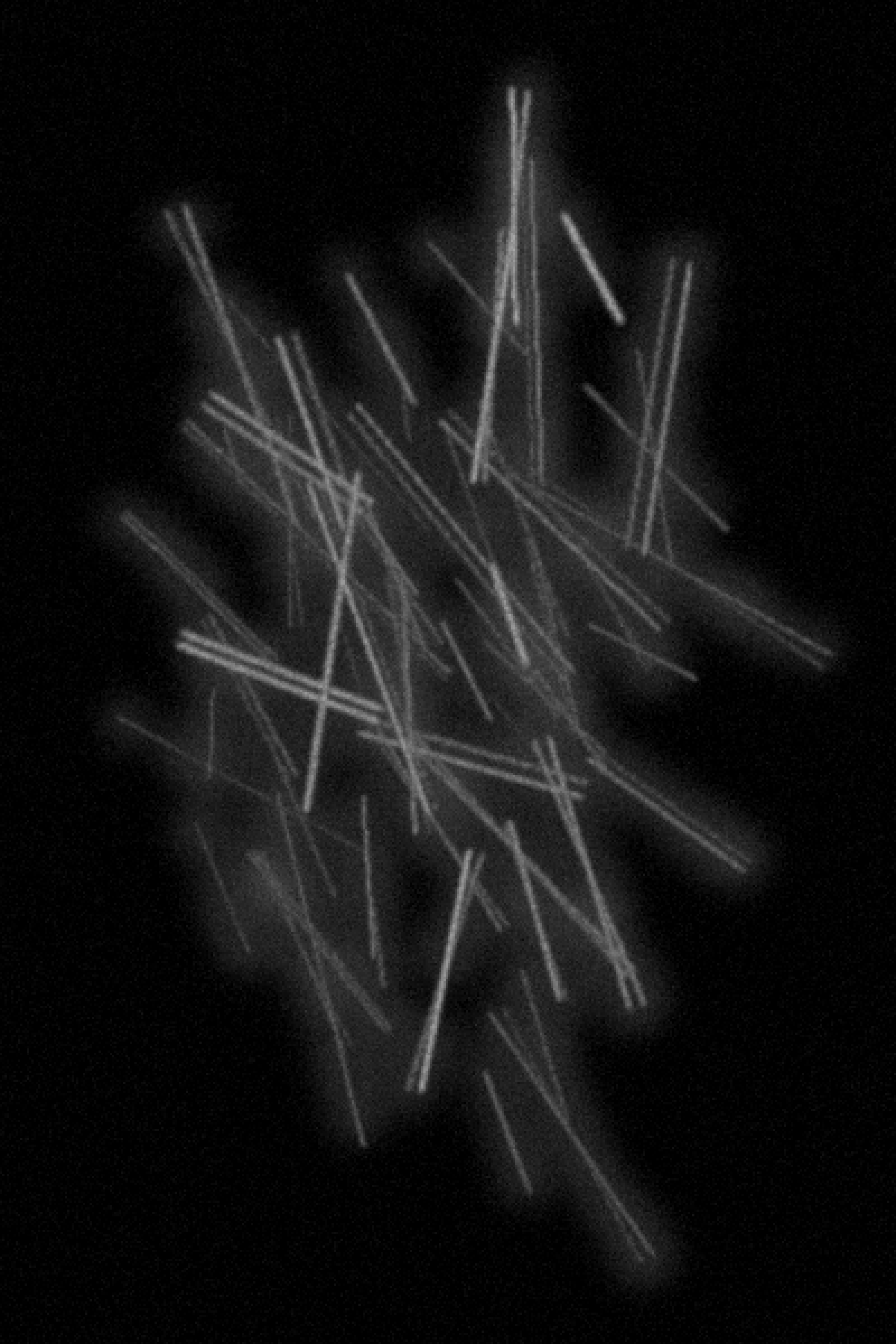}
  \caption{{\bf Simulated test case image} featuring many parallel lines at different distances, some very close. It also contains several crossings of lines of different brightness at small angles. The image is used as a test case to qualitatively highlight the performance of the three methods compared here in terms of filament pixel detection and especially structure detection. As the FS is the only method that extracts line data, the notion of parallel or intersecting lines only makes sense for this method.}
  \label{test_case_cell}
\end{figure}

To illustrate the limits of the methods compared qualitatively, in view of the challenges outlined as IIa) to IIc) from Section ``The Filament Sensor and the Benchmark Dataset'', we picked three suitable examples from the benchmark dataset and the specifically simulated image, Fig. \ref{test_case_cell}. In this context, ``segmentation'' means the detection of lines or line pixels. Therefore we refer to the detection of excess lines or line pixels as oversegmentation.

\begin{figure}[!ht]
  \centering
  \includegraphics[height=0.7\textheight]{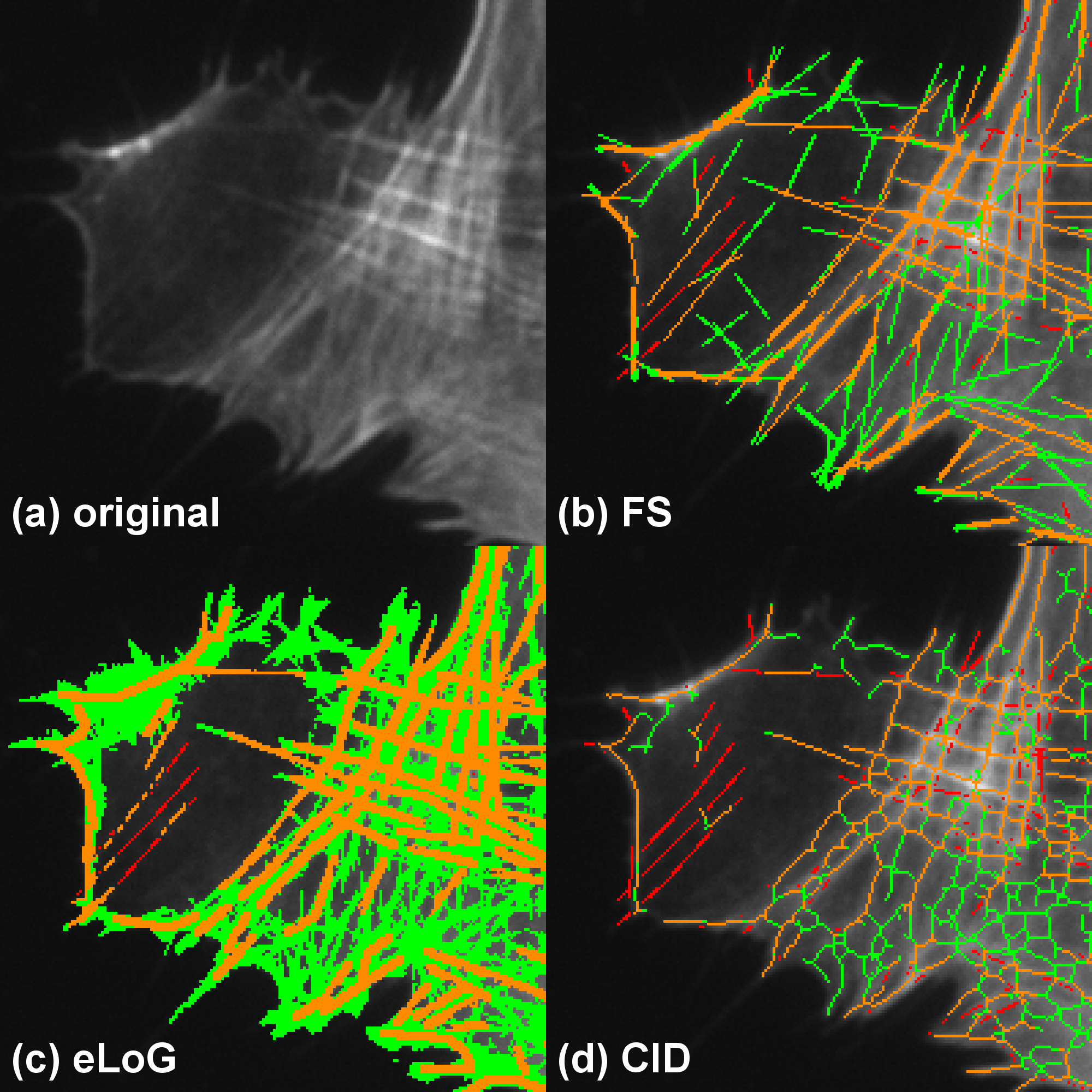}
  \caption{{\bf Performance comparison for inhomogeneous brightness and crossing lines.} Showing a detail of cell M3. The subfigures represent (a) the original detail, (b) the results of the FS, (c) the results of the eLoG method and (d) the results of CID. Green pixels are false positives detected by the method, yellow are correctly identified pixels and red are missed pixels as in Fig. \ref{pixels_image_real}.
  \\
  The FS produces a fair amount of false positives but fares quite well both in the dark region on the left as well as the bright region with crossing lines on the right. The eLoG method also find parts of the lines in the dark region albeit at the expense of significant oversegmentation in the bright region. CID detects lines almost exclusively in the higher contrast bright region, where it produces a cobweb structure with an amount of oversegmentation similar to the FS.
  }
  \label{qualitative_inhom}
\end{figure}

\paragraph{Comparison for inhomogeneous brightness and crossing of lines} is illustrated in Fig. \ref{qualitative_inhom} which shows a detail from cell image M3. The upper right image regions display crossings of filaments which are almost completely captured by the FS and slightly less by CID, that tends to produce a network structure with only short straight segments. The eLoG method highly oversegments, essentially identifying all pixels in this region. Oversegmentation is also done by the FS and CID, but on a much lower scale. Image inhomogeneity is introduced by the dark area on the left. Clearly, the FS is the only method that finds most of the labeled segments, followed by the eLoG method that tends to break longer lines into pieces. While CID finds almost none of the filaments, together with the eLoG method it also features almost no oversegmentation in this area. In contrast, the oversegmentation by the FS features line segments that are visible in the raw image also, that were not labeled by the human expert.

\begin{figure}[!ht]
  \includegraphics[width=\textwidth]{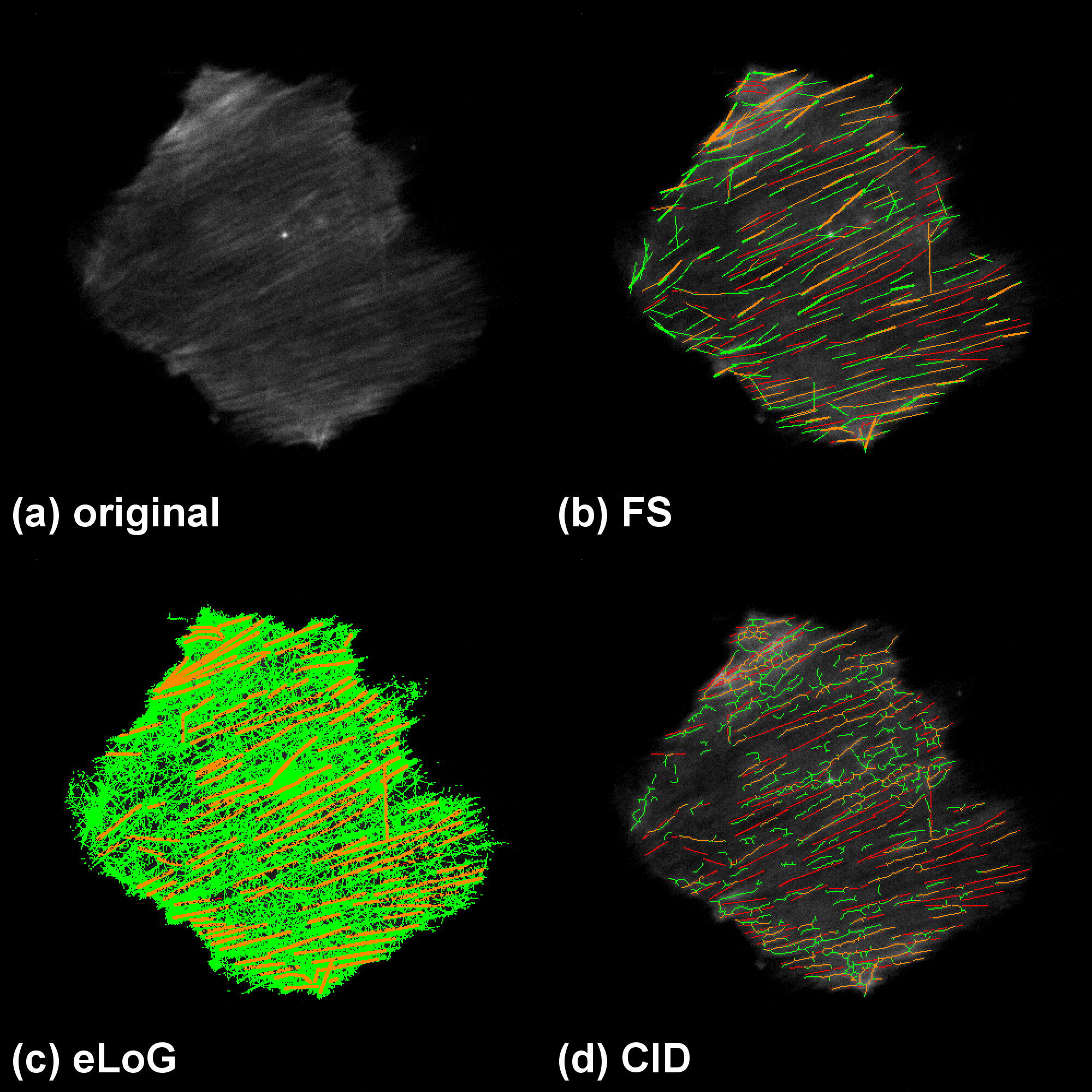}
  \caption{{\bf Performance comparison in the presence of blur.} Showing cell VB2. Green pixels are false positives detected by the method, yellow are correctly identified pixels and red are missed pixels as in Fig. \ref{pixels_image_real}. All methods produce some false positives but the eLoG method stands out by detecting almost all cell pixels as line pixels. CID produces a cobweb structure with a similar amount of oversegmentation as the FS.}
  \label{qualitative_blur}
\end{figure}

\begin{figure}[!ht]
  \includegraphics[width=\textwidth]{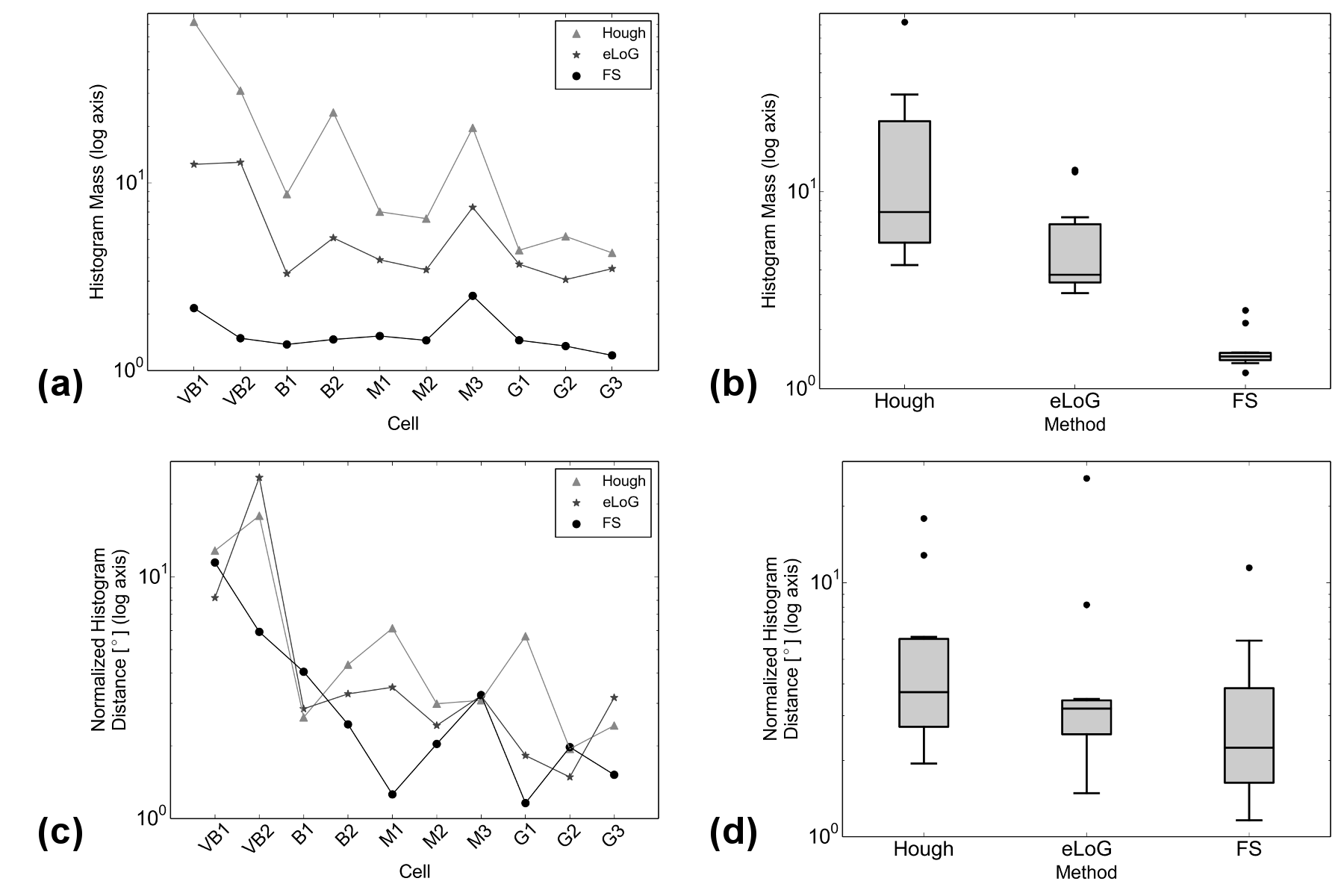}
  \caption{{\bf Comparison of angular histograms.} Log-scales and logs of earth movers distances are compared independently. Subfigure (a) shows the log-scaled relative mass of the histograms with respect to the histogram from filaments marked by a human expert for individual cells. In subfigure (b) boxplots of the log-scaled relative masses each over all $10$ cells are displayed. Subfigure (c) shows the logs of earth movers distance between normalized histograms of the evaluation method and the ground truth; subfigure (d) displays the corresponding boxplots each over all $10$ cells. The plots are semi-logarithmic because scales vary widely. Data points corresponding to same methods in plots (a) and (c) are connected only for better visualization.}
  \label{angle_eval_real}
\end{figure}

\begin{figure}[!ht]
  \includegraphics[width=\textwidth]{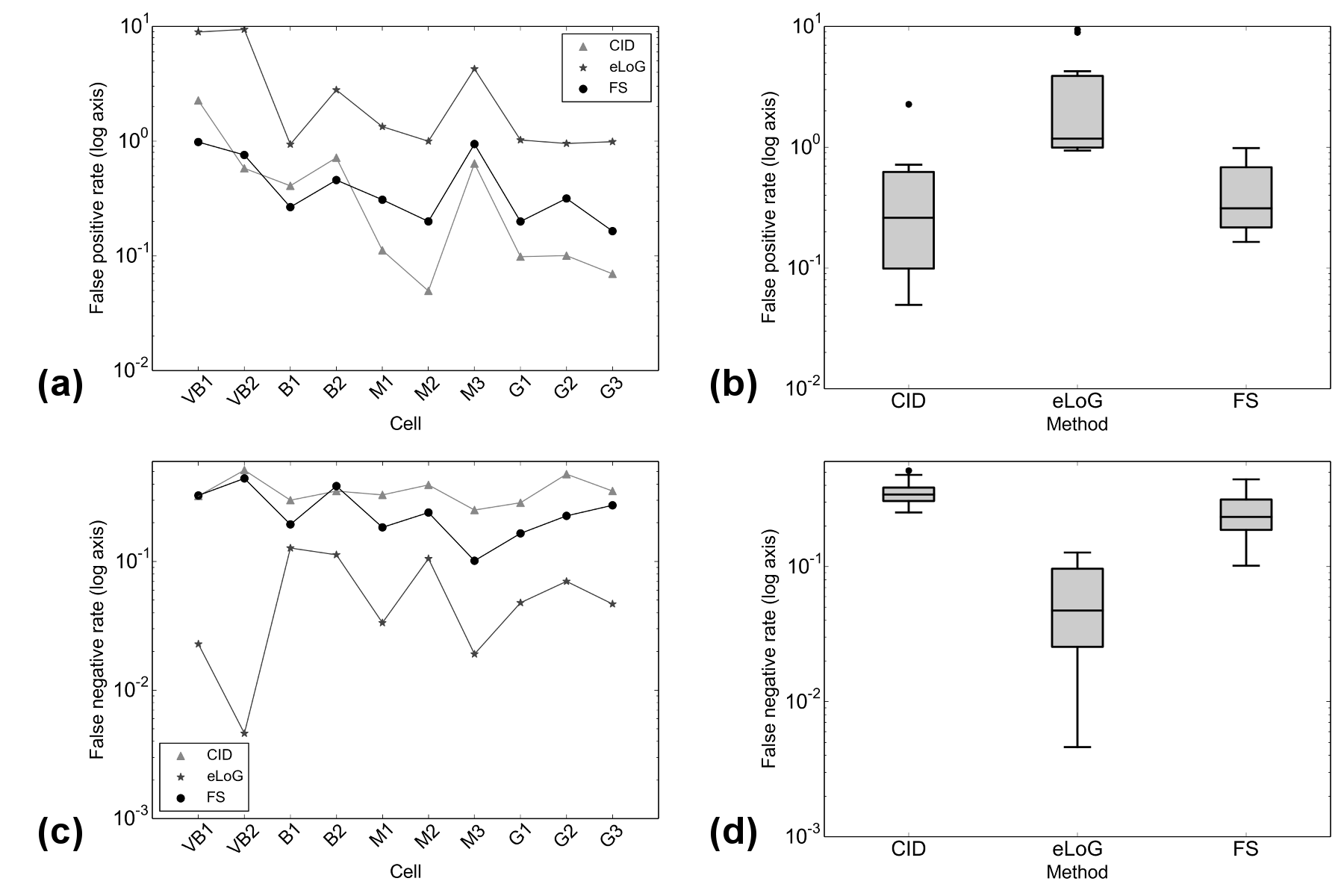}
  \caption{{\bf Comparison of $R_{\textnormal{fp}}$ (false positive ratios) and $R_{\textnormal{fn}}$ (false negative ratios).} Subfigure (a) displays $R_{\textnormal{fp}}$ for individual cells, subfigure (b) the corresponding boxplots over all $10$ cells. Subfigure (c) displays $R_{\textnormal{fp}}$ for individual cells, subfigure (d) shows the corresponding boxplots over all $10$ cells. The plots are semi-logarithmic because scales vary widely. Data points corresponding to same methods in plots (a) and (c) are connected only for better visualization.}
  \label{pixel_eval_real}
\end{figure}

\paragraph{In the presence of blur,} e.g. if the image is slightly off-focus, cf. Fig. \ref{qualitative_blur} showing cell VB2, the FS identifies $55.74\, \%$ of labeled filament pixels with an oversegmentation of $75.83\,\%$ of labeled filament pixels. The CID finds $48.65 \,\%$ of labeled filament pixels with an oversegmentation rate of $58.02\,\%$. Of course, it cannot give orientation information. Notably the orientation due to oversegmentation of the FS is compatible with the ground truth orientation labeling. The eLoG method dramatically oversegments, rendering its result useless for further analysis. This cell VB2 is one of the outliers both for the eLoG and Hough methods, cf. Figs. \ref{angle_eval_real} and \ref{pixel_eval_real}.

\begin{figure}[!ht]
  \includegraphics[width=\textwidth]{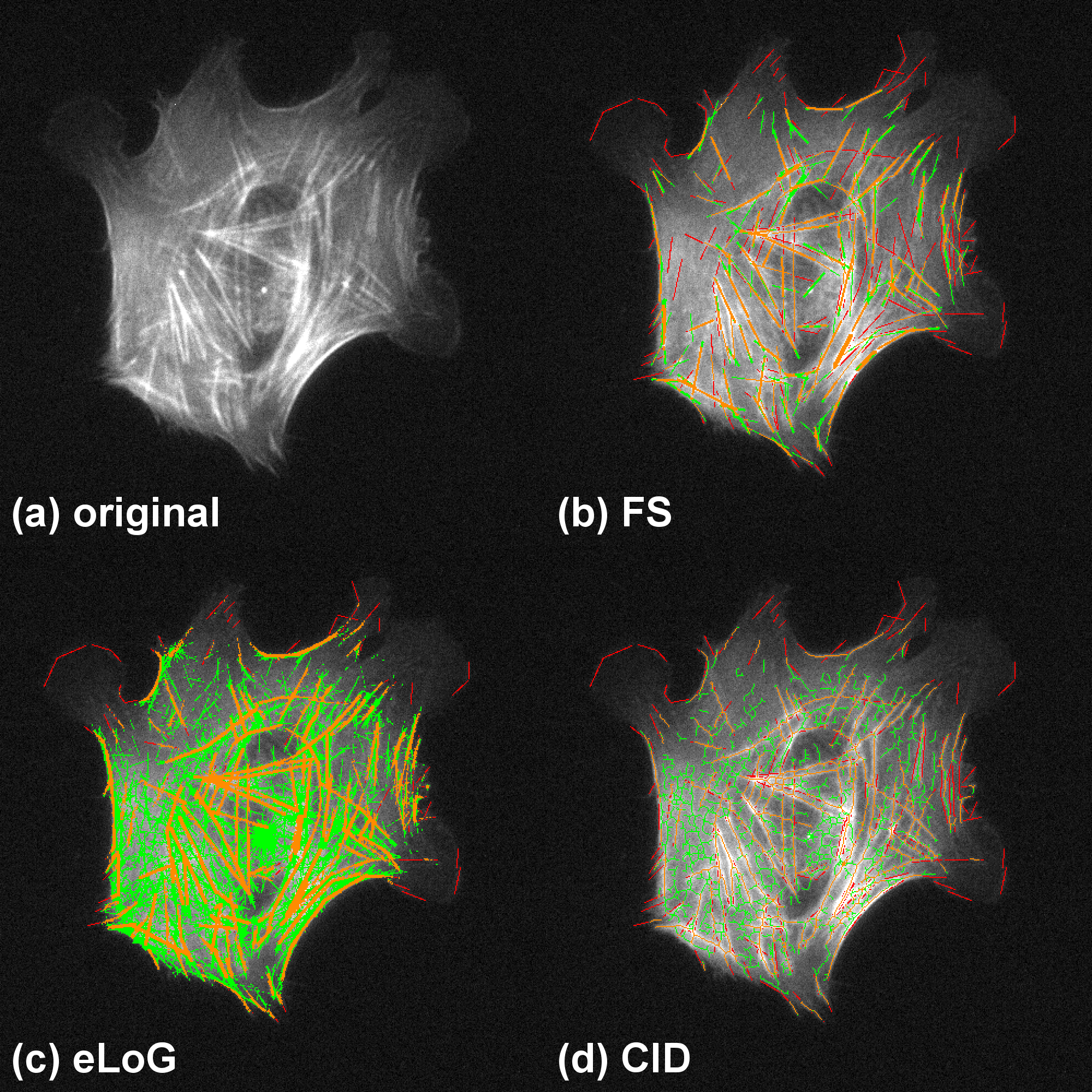}
  \caption{{\bf Performance comparison in the presence of noise.} Showing cell B2. Green pixels are false positives detected by the method, yellow are correctly identified pixels and red are missed pixels as in Fig. \ref{pixels_image_real}. For this image the FS fares much better than both other methods. The eLoG method as well as CID both find a large amount of spurious features, where the eLoG method detects large contiguous areas and CID produces a cobweb structure, covering nearly the whole cell. Especially in the left, lower, and central parts of the image the FS is the only method that does not detect a large amount of spurious line pixels.}
  \label{qualitative_noise}
\end{figure}

\paragraph{Under noise} as in Fig. \ref{qualitative_noise} showing cell B2, the FS identifies $61.40\, \%$ of labeled filament pixels with oversegmentation rate of $45 \,\%$ while CID finds $64.73 \,\%$ with an oversegmentation rate of $71.75\,\%$. Again due to heavy oversegmentation, the eLoG method's results cannot be used for further analysis.

\begin{figure}[!ht]
  \centering
  \includegraphics[height=0.7\textheight]{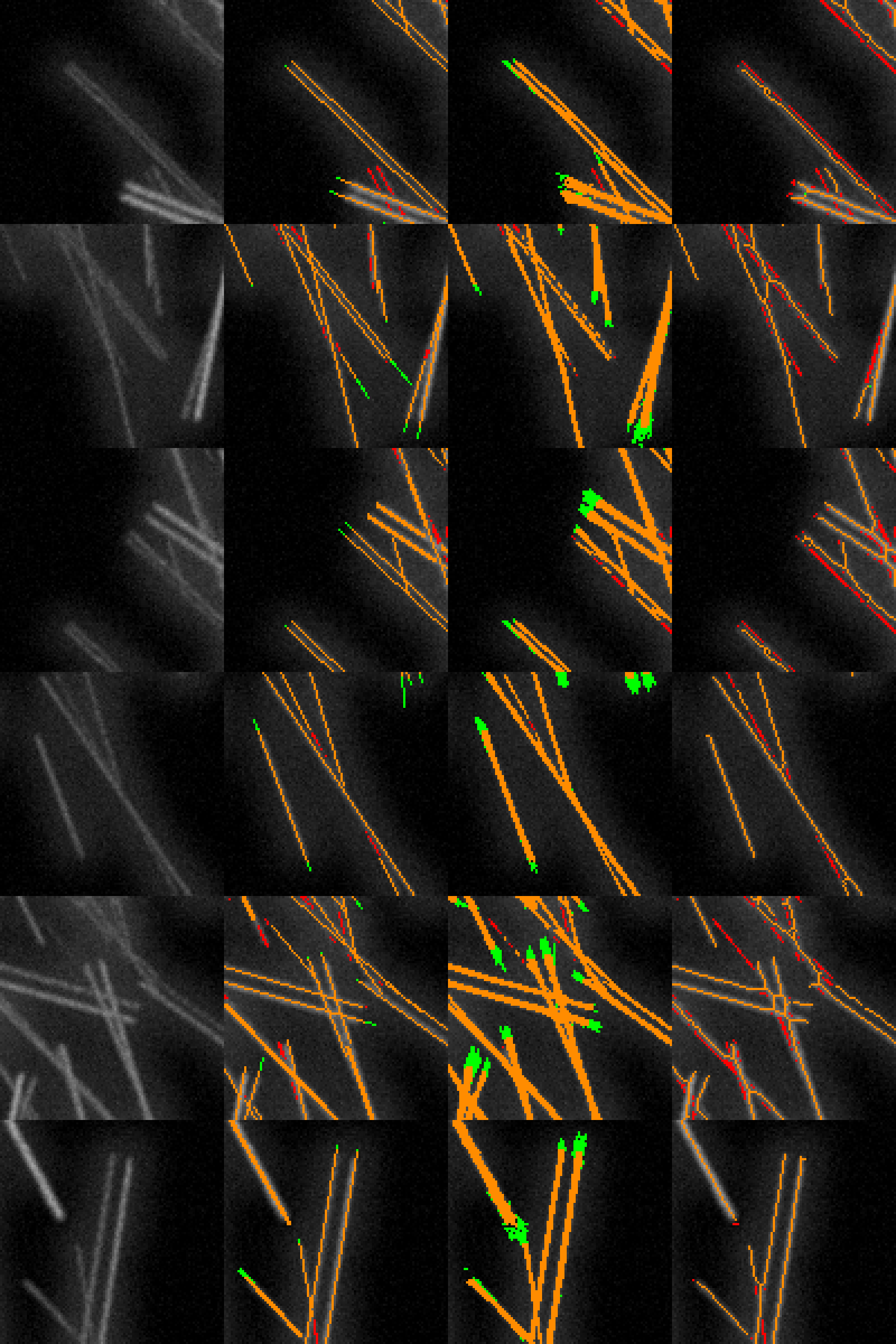}
  \caption{{\bf Performance comparison for close parallel lines and lines crossing at small angles.} Showing details from the simulated test image displayed in Fig. \ref{test_case_cell}. The columns show, from left to right, the original detail, the results of the FS, the results of the eLoG method, and the results of CID. Green pixels are false positives detected by the method, yellow are correctly identified pixels and red are missed pixels as in Fig. \ref{pixels_image_real}. In the details shown the FS and eLoG method clearly outperform CID, which does poorly at detecting close parallel lines, as shown in the first three rows of the figure. Also for crossing lines as displayed in the fourth and fifth row, CID performs poorly. The eLoG method shows some weaknesses for close parallel lines and overshoots at the endings of lines. Overall, its line pixel detection capabilities in these synthetic test cases are comparable to those of the FS.}
  \label{qualitative_sim}
\end{figure}

\paragraph{Parallel lines and small angles.}
Finally, Fig. \ref{qualitative_sim} displays details from the test image displayed in Fig. \ref{test_case_cell}. Close parallel lines of three pixel distance (the narrow lines in the first three rows) are well identified by the FS but not classified as two lines by CID and often only as one line with the other broken into small pieces (or introducing pieces connecting the two lines) by the eLoG method. For smaller distances (one and two pixels) in the fourth row, also the FS fails.

In penultimate row of Fig. \ref{qualitative_sim}, angles up 6.5$^\circ$ (two-tined fork in the centers of the penultimate row) are well resolved by FS and the eLoG method, CID fails, however. Lower angles (3.5$^\circ$, located southwest of the fork in the same row) are not resolved by any of the methods. While the angle of 5$^\circ$ (in the center of the fourth row images) is almost fully resolved only by the eLoG method, we stress that the eLoG method only identifies pixels; extracting line information would require separate tracing, with the segment sensor, say, which tends to destroy this angular resolution.

Notably, the eLoG method tends to make lines thicker and longer, cf. the bottom row of Fig. \ref{qualitative_sim}.

\paragraph{The nature of oversegmentation} is distinctively different among the three methods. All methods tend to oversegment in challenging images. The eLoG method tends to solidly fill entire image areas and CID tends to produce fine network structures with straight lines only at very small scales. The FS produces false straight segments, often parallel to labeled filaments. 

\paragraph{Challenges summary.} While the eLoG method tends to make lines thicker and longer and under noise and blur to fill the entire image, CID tends to return a network structure, performing poorly at close parallel lines and lines intersecting with small angles. With the eLoG method it shares the difficulty to identify features under image inhomogeneity. Regarding image inhomogeneity, noise and (nearly) parallel line detection, the FS outperforms the other methods. Regarding line crossings, the eLoG method seemingly outperforms the FS which outperforms CID, however only ``seemingly'' because the eLoG method would have to be followed by a tracing step. Under blur the FS features more false positives than CID, the FS's false positives tend to reflect labeled nearby filament orientation better than those of CID.

Overall, the FS performs well in terms of the challenges lined out before and often outperforms the eLoG and CID methods in terms of correct line pixel identification.

\subsection*{Angular Histograms}
\begin{figure}[!ht]
  \includegraphics[width=\textwidth]{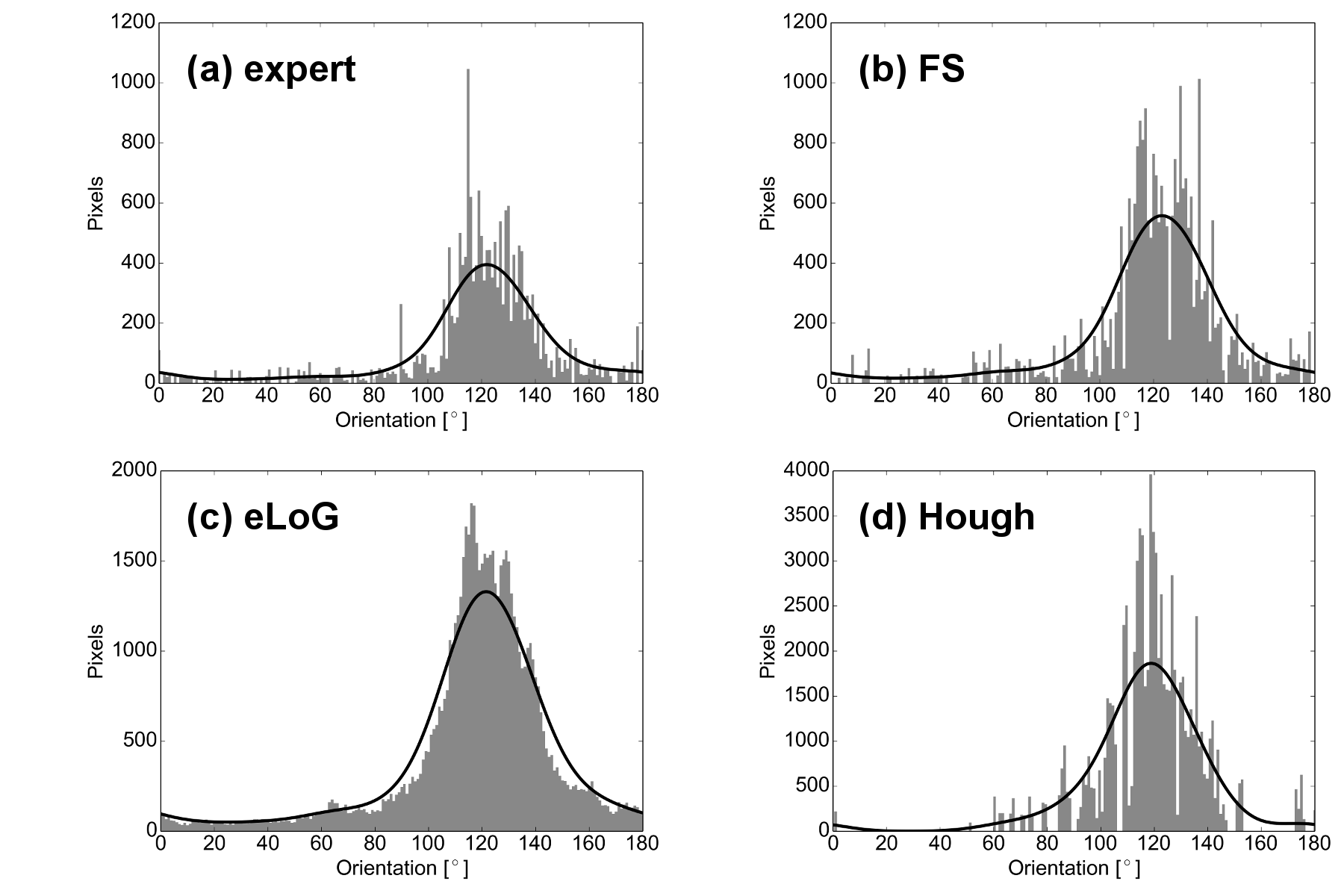}
  \caption{{\bf Angular histograms of filament mass for an hMSC.} The cell G1 used for this illustration is a fixed cell that has been immunofluorescently stained. The subfigures display: (a)~ground~truth, (b)~FS~results, (c)~eLoG~results, (d)~Hough~results. The black curves illustrate the result of kernel smoothing with a Gaussian of standard deviation $\sigma = 10$.}
  \label{hist_real}
\end{figure}

\begin{figure}[!ht]
  \includegraphics[width=\textwidth]{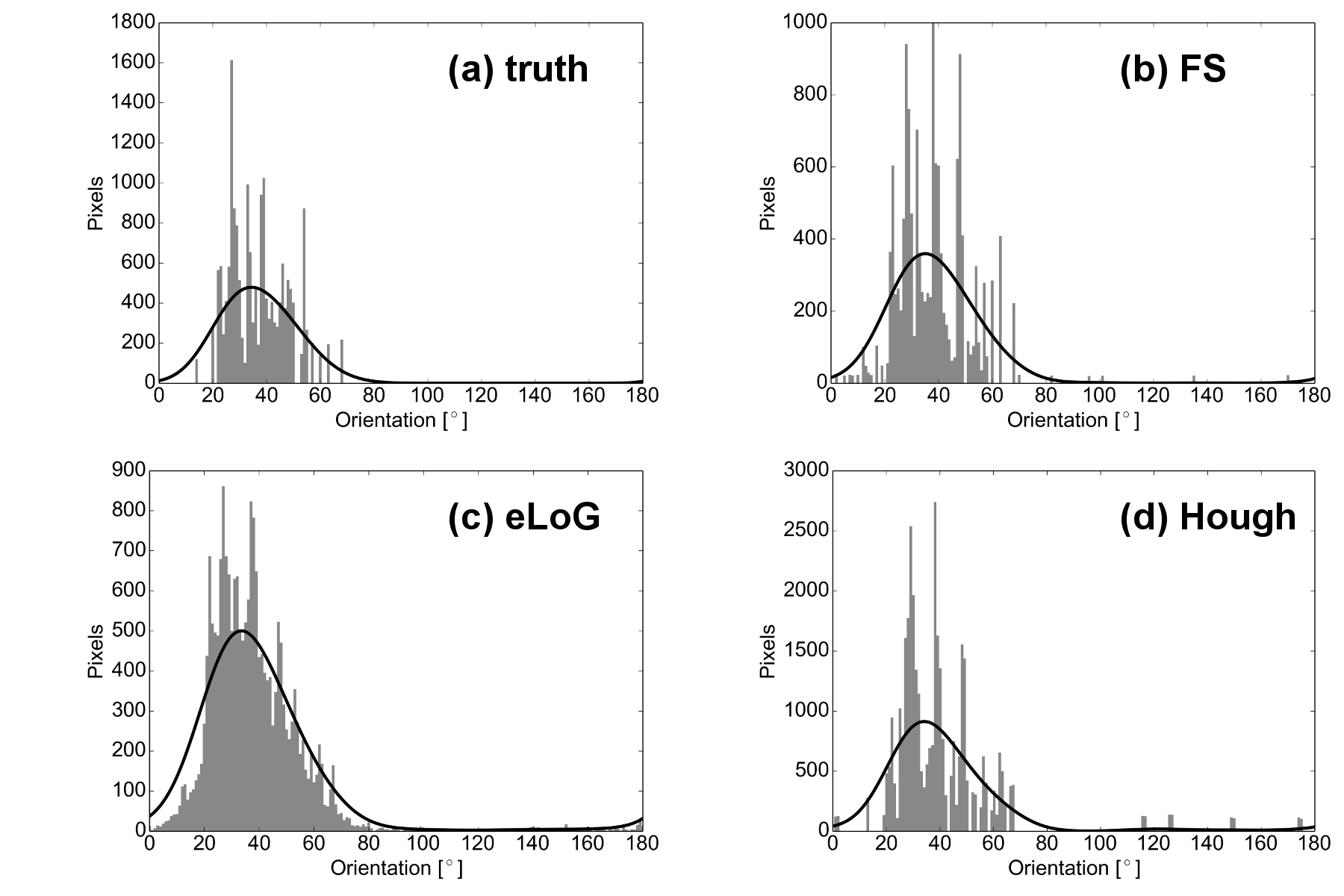}
  \caption{{\bf Angular histograms of filament mass for a simulated cell} (cell 05). Subfigures and black line are as in Fig. \ref{hist_real}.}
  \label{hist_sim}
\end{figure}

As noted before, the two methods eLoG and HT allow for angular histograms (accumulated filament length per angular orientation) as displayed as well for the FS in Figs. \ref{hist_real} and \ref{hist_sim}.

\begin{figure}[!ht]
  \includegraphics[width=\textwidth]{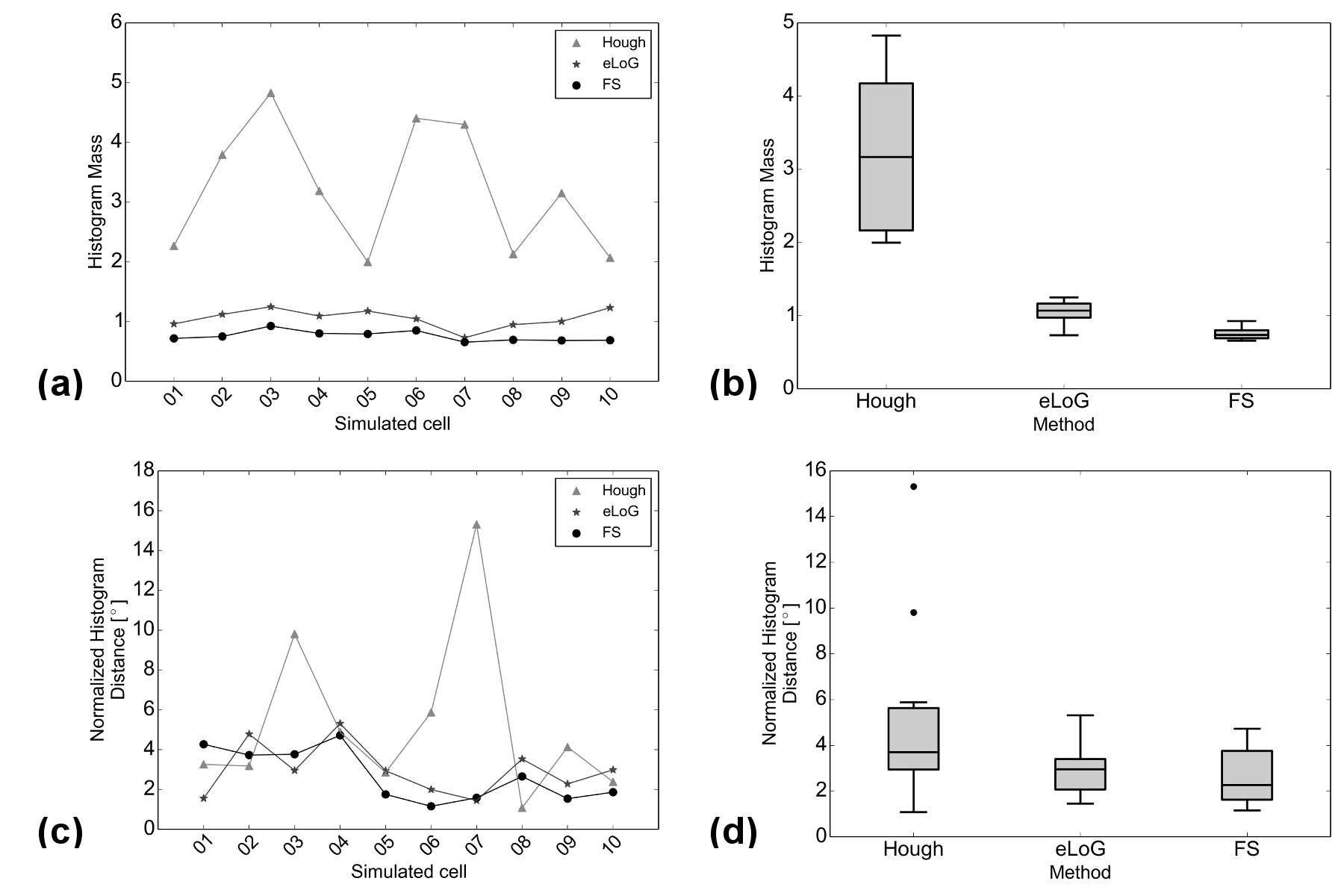}
  \caption{{\bf Comparison of angular histograms.} Subfigures are as in Fig. \ref{angle_eval_real}, now for simulated cells, with all relative masses and distances now linear. Data points corresponding to same methods in plots (a) and (c) are connected only for better visualization.}
  \label{angle_eval_sim}
\end{figure}

\paragraph{Total histogram mass.}
Inspecting the ratio of detected total mass of histograms over total mass of ground truth histograms (top rows of Figs. \ref{angle_eval_real} and \ref{angle_eval_sim}) we observe the following.

\begin{figure}[!ht]
  \includegraphics[width=\textwidth]{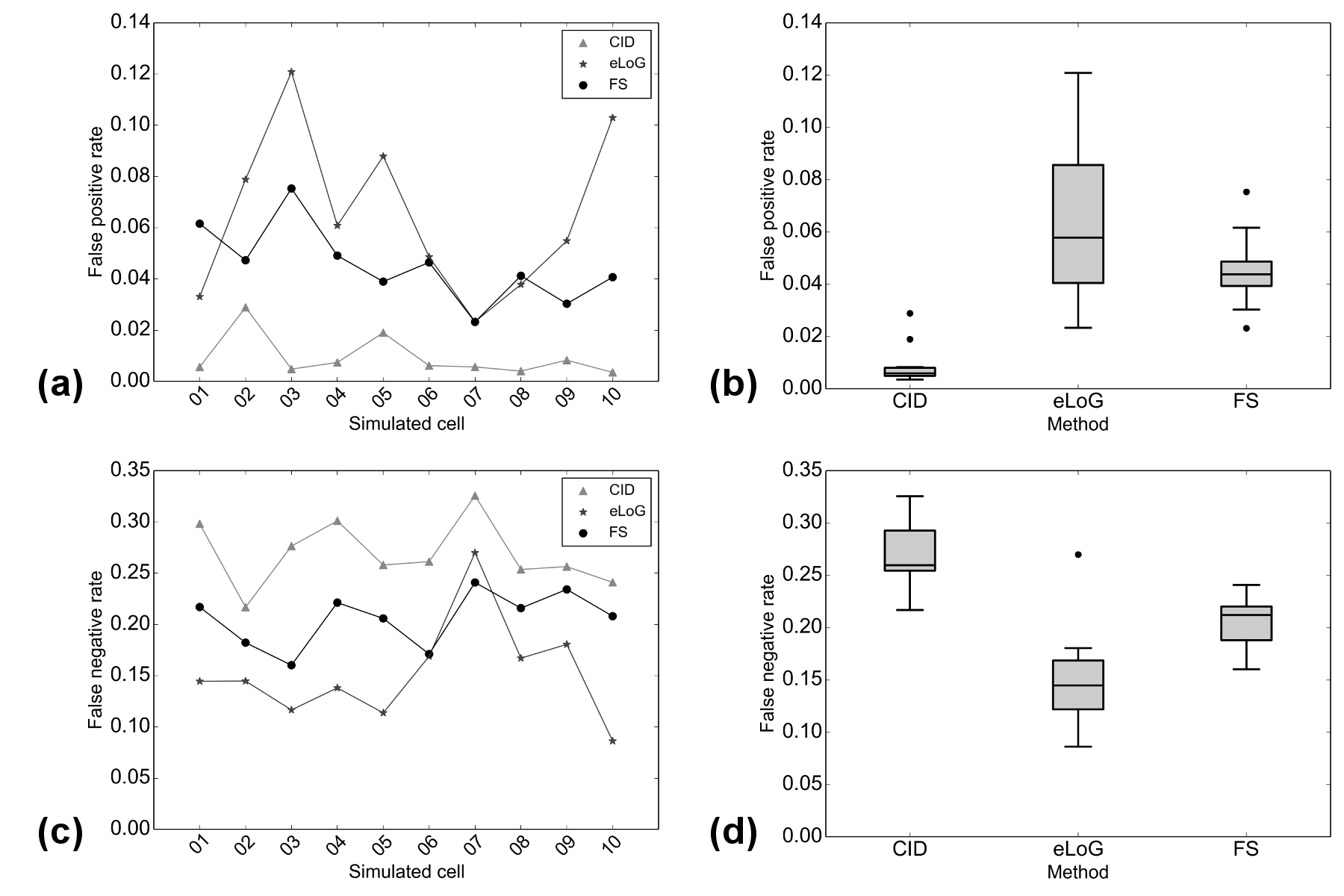}
  \caption{{\bf Comparison of $R_{\textnormal{fp}}$ (false positive ratios) and $R_{\textnormal{fn}}$ (false negative ratios).} Subfigures are as in \ref{pixel_eval_real}, now for simulated cells, with all axes linear. Data points corresponding to same methods in plots (a) and (c) are connected only for better visualization.}
  \label{pixel_eval_sim}
\end{figure}

\begin{enumerate}
  \item For real cells, all three methods detect too much mass, the amount of excess mass correlates negatively with image quality. This effect is consistently strongest for HT (the maximum is 72.2--fold mass) and weakest for the FS (between 1.2-- and 3--fold mass) while the eLoG method yields results in between (between 3-- and 12.9--fold mass). The good performance of the FS is a consequence of the usage of the expert knowledge that blurred lines appear broadened in the binary image. By their very construction this expert knowledge cannot be incorporated by the alternate methods.
  \item For simulated cells, HT detects consistently slightly too much mass (between 1.3-- and 2.8--fold) while eLoG and FS detect consistently too little mass (between 0.54-- and 1.00--fold).
\end{enumerate}

\paragraph{Normalized histograms} are obtained by dividing by total histogram mass, making them comparable to one another, and in particular to ground truth normalized histograms. For comparison we use the shortlist implementation \cite{GottschlichSchuhmacher2014} of the earth mover's (or 1st Wasserstein) distance of histograms with arc length as ground distance. Because it relates masses in different bins in a natural way, this distance measure is well suited for comparison of histograms as shown by \cite{RubnerTomasiGuibas2000}. We normalize histograms, because the canonical computation of the EMD requires both histograms to have the same mass. Note that this comparison disregards the effect of excess mass on shape features, the noise level, say. For a complete picture with ground truth available, however, from the illustration in the bottom rows of Figs. \ref{angle_eval_real} and \ref{angle_eval_sim} we observe that
\begin{enumerate}
  \item on the average the FS yields the lowest distance, directly followed by eLoG, where the FS outperforms HT in most of the cases;
  \item the eLoG method performs slightly better than the FS on smaller cells with small ground truth mass; those sparse histograms are very sensitive to detection errors, an effect which is damped by the large excess mass the eLoG method produces;
\end{enumerate}
In consequence the FS is (on the average) the most robust even in terms of histogram shape only, ignoring excess mass.

\subsection*{Pixel Identification}

The CID method does not aim at identifying filament orientations, furthermore, it only returns pixels along thin filament skeletons having width one. For comparison in terms of correctly identified filament pixels we thus define the following procedure.

Let $N_{\textnormal{t}}$ denote the number of ground truth pixels for which at least one pixel identified by the method is in a $3\times3$-square around it, $N_{\textnormal{fn}}$ the number of other ground truth pixels and $N_{\textnormal{fp}}$ the number of pixels detected by the method for which no ground truth pixel is in a $3\times3$-square around it. We define the false negative ratio as $R_{\textnormal{fn}} = N_{\textnormal{fn}} / N$ where $N= N_{\textnormal{t}} + N_{\textnormal{fn}}$ and the false positive ratio as $R_{\textnormal{fp}} = N_{\textnormal{fp}} / N$. The results are displayed in Figs. \ref{pixel_eval_real} and \ref{pixel_eval_sim}.

This comparison which can also be performed for eLoG shows that 
\begin{enumerate}
  \item for real cells, in terms of detecting true filament segments, consistently, eLoG performs best, followed by the FS, which is closely followed by CID; this effect seems not to correlate with image quality; 
  \item for real cells, in terms of non-detection of non-marked filament segments, consistently, CID performs best, closely (note the logarithmic scale) followed by the FS and further followed by eLoG; this effect -- which is much stronger than the one preceding -- correlates negatively with image quality and once again illustrates the tendency of the eLoG method toward detecting consistently far too many filament pixels;
  \item for simulated cells, similar effects are visible, yet on a much smaller scale, however, in terms of missing true segments, the FS outperforms CID and in terms of over-detection, the FS performs equally less optimal as eLoG, where the absolute rates are very low for all methods. 
\end{enumerate}

In summary, CID finds too little pixels and the eLoG finds too many while the FS achieves the balance between detecting too many and too little filament pixels.

\subsection*{Speed}
Using an orientational grid of $180$ angles, eLoG takes more than $20$ minutes per image compared to approximately $20$ seconds for the FS and for HT (these runtimes have been observed on a single core 1.73 GHz Intel Celeron with 2 GB ram). In contrast, the CID method required more than 20 hours (running on an AMD Opteron 6140 with 32 cores at 2.6 GHz and 128 GB ram). The disparity of runtimes is illustrated in Table \ref{table}.

The comparatively short runtime in connection with the semi-automated nature of its workflow (fast, uninterrupted batch processing) thus makes the FS a viable tool for analyzing the actin cytoskeleton via time series of live cell images. Especially, as $50$ images are taken every $10$ minutes in our setup, the FS can analyze the full filament structure of images almost in real time. This cannot be achieved with any of the existing methods. A future application in terms of automatic microscopy and real time stress fiber analysis is thus conceivable.

\section*{Discussion}

We have developed the Filament Sensor (FS) that allows for nearly real time and semi-automated extraction of straight filament structures in microscope images, in particular from live human mesenchymal stem cells (hMSCs). Reliable extraction of the entire filament data is essential for a better understanding and future modelling of the complex mechano-sensing phenomena e.g. for detailed statistical studies of the relationship between matrix elasticity and early mechano-guided differentiation of hMSCs or myoblast differentiation \cite{EnglerSenSweeneyDisher2006,Zemel2010a,Yoshikawa2013}. In view of live cell imaging, approximately 4000 images (30 cells followed over 24 hours with an image every 10 minutes) need to be processed per day. To the knowledge of the authors, no method has previously been available for this task (even without extracting the entire filament structure, the conventional eLoG method would require 60 days for this, CID almost 10 years).

We have provided for a proof of concept by checking against a database of filaments manually marked by a human expert and a simulated database. Moreover, we have compared our FS against three other methods, the output of which, however, encompasses only a small portion of the entire filament data we are interested in. In this comparison, the FS performs best in terms of histogram mass and histogram distances and takes a middle ground between false positives (where, for real cells, eLoG is best and CID is worst) and false negatives (where, for real cells, CID is best and eLoG is worst).

In terms of speed, the FS overwhelmingly outperforms all competitors. While the HT would require additional tracing, eLoGs are slower by a factor of more than $60$ and CID by a factor of more than $10^3$ (on a stronger machine). Clearly the latter is beyond any acceptable runtime for the processing of thousands of images we are aiming at.

In summary, recalling goals I), II) and III) from Section ``The Filament Sensor and the Benchmark Dataset'', the FS is the first tool available to extract complete filament data from hMSCs images (goal III), unsupervised in near real time (goal I) that is equally or more robust (goal II) than slower competitors with more limited output.

At the end of this exposition we would like to point out two challenges we leave for future work.
\begin{enumerate}
  \item As the FS starts finding wide lines first and then proceeds to thinner lines, it will sometimes detect lines of variable width as several pieces with different width. These line fragments could be matched to produce a single long line. 
  \item A future version of the FS may include slightly curved stress fibers and a correspondingly appropriate matching procedure. We plan to explore the application of curved Gabor filters \cite{Gottschlich2012} for detecting curved stress fibers and for coping with fiber crossings. Recently, Gabor wavelets have proved to be beneficial for a similar application in retinal vessel tracking \cite{BekkersDuitsBerendschotTerhaarromeny2014}.
\end{enumerate}


\section*{Acknowledgments}

All authors gratefully acknowledge funding by the Deutsche Forschungsgemeinschaft (DFG) within the collaborative research center SFB 755 ``Nanoscale Photonic Imaging' project B8 and the Open Access Publication Funds of the University of G\"ottingen. C. Gottschlich and S. Huckemann also gratefully acknowledge support of the Felix-Bernstein-Institute for Mathematical Statistics in the Biosciences and the Niedersachsen Vorab of the Volkswagen Foundation. The authors acknowledge Julian R\"{u}ger's contribution to an earlier version of the FS. The authors gratefully acknowledge the detailed remarks of three anonymous reviewers and the editor that have greatly contributed to improving this article.


%
%
%

\end{document}